\documentclass[
  journal=medium,
  manuscript=article,
  year=2020,
  volume=37,
]{cup-journal}

\usepackage[utf8x]{inputenc}
\usepackage{amsmath}
\usepackage[nopatch]{microtype}
\usepackage{booktabs}

\usepackage[english]{babel}
\usepackage{graphicx}
\usepackage[colorinlistoftodos]{todonotes}
\usepackage{array}
\usepackage{siunitx}
\usepackage[hidelinks]{hyperref}       
\usepackage{url}            
\usepackage{amsfonts}       
\usepackage{nicefrac}       
\usepackage{multicol}
\usepackage{tcolorbox}
\usepackage{geometry}
\usepackage{svg}
\usepackage{tabularx}
\usepackage[export]{adjustbox}
\usepackage{booktabs}
\usepackage{amssymb}
\usepackage{caption}
\usepackage{subcaption}
\usepackage{multirow}
\usepackage{algorithm}
\usepackage{algorithmic}
\usepackage{enumitem} 
\usepackage{tabularx}
\usepackage{amsmath}
\usepackage{mathtools}
\usepackage{float}
\usepackage{soul} 
\usepackage{xcolor} 

\usepackage[]{biblatex-chicago}

\DeclareMathOperator*{\argmax}{arg\,max}

\setlist[itemize]{leftmargin=*}

\newcolumntype{H}{>{\setbox0=\hbox\bgroup}c<{\egroup}@{}}

\title{Enhancing LLM-Based Text Classification in Political Science: Automatic Prompt Optimization and Dynamic Exemplar Selection for Few-Shot Learning}

\author{Menglin Liu}
\affiliation{University of California, Davis}
\email[F. Author]{mliliu@ucdavis.edu}

\author{Ge Shi}
\alsoaffiliation{University of California, Davis}
\email[]{geshi@ucdavis.edu}

\keywords{text classification, sentiment analysis, stance detection, in-context learning, automatic prompt optimization, large language models (LLMs)} 

\begin{document}

\maketitle

\begin{abstract}

Large language models (LLMs) offer substantial promise for text classification in political science, yet their effectiveness often depends on high-quality prompts and exemplars. To address this, we introduce a three-stage framework that enhances LLM performance through automatic prompt optimization, dynamic exemplar selection, and a consensus mechanism. Our approach automates prompt refinement using task-specific exemplars, eliminating speculative trial-and-error adjustments and producing structured prompts aligned with human-defined criteria. In the second stage, we dynamically select the most relevant exemplars, ensuring contextually appropriate guidance for each query. Finally, our consensus mechanism mimics the role of multiple human coders for a single task, combining outputs from LLMs to achieve high reliability and consistency at a reduced cost. Evaluated across tasks including sentiment analysis, stance detection, and campaign ad tone classification, our method enhances classification accuracy without requiring task-specific model retraining or extensive manual adjustments to prompts. This framework not only boosts accuracy, interpretability and transparency but also provides a cost-effective, scalable solution tailored to political science applications. An open-source Python package (PoliPrompt) is available on GitHub.
\end{abstract}
\section{Introduction}

Text data is a key resource in political science, where text classification tasks—such as stance detection, sentiment analysis, and topic classification—are widely used to study political discourse and public opinion. For example, sentiment analysis of campaign ads can track trends in messaging tone, helping researchers understand their impact on voter perception (\cite{fowler2021political}). In authoritarian contexts, text classification sheds light on public opinion about government actions, as demonstrated by \citet{pan2018concealing}, who categorize online posts in China by support or criticism of local governance, and by \citet{king2013censorship}, who analyze censorship patterns in Chinese social media. Text classification also plays an essential role in legislative studies. \citet{grimmer2013text}) show how categorizing legislative texts and party manifestos helps quantify ideological shifts over time, while \citet{slapin2008scaling} develop a model to track party positions by analyzing speeches in parliamentary debates. On social media, classification has been used to examine political polarization and public engagement. \citet{bond2015quantifying} analyze social media topics to assess voter engagement, and \citet{barbera2015understanding} uses Twitter data to reveal ideological clustering. Collectively, these studies demonstrate how text classification is a common method in political science for analyzing complex data and understanding political behavior.

Political scientists have long relied on \textbf{training-based methods} for text classification, including Statistical Language Models (SLMs) and, more recently, Pre-trained Language Models (PLMs) (\cite{8950616, 9079815, 1174243, devlin2019bertpretrainingdeepbidirectional, sun2020finetuneberttextclassification}). SLMs, often paired with classifiers like Support Vector Machines (SVMs), capture basic word patterns but require extensive manual feature engineering, making them labor-intensive. Importantly, SLMs are highly task-specific: a model trained for one purpose, such as topic classification, must be retrained for a different task like sentiment analysis or for application on new types of text, such as social media posts versus news articles. PLMs, like BERT, offer advancements by generating context-aware word representations that require less manual feature engineering and capture more subtle relationships between words.

In recent years, Large Language Models (LLMs) have revolutionized text classification by enabling flexible and efficient learning without extensive task-specific training. A key mechanism driving this capability is \textbf{in-context learning} (ICL), where LLMs perform classification using only a well-crafted prompt, eliminating the need for extensive task-specific training or fine-tuning on labeled data.\footnote{A prompt typically consists of four main components: (1) task description, which provides context for the model, (2) few-shot examples, which guide the model’s responses based on prior cases, (3) the query, where the model generates predictions based on the given input question or context, and (4) the prefix, which serves as an additional guiding phrase or instruction to shape the model's output format (\cite{brown2020languagemodelsfewshotlearners}).} \footnote{ A detailed comparison between in-context learning and fine-tuning is presented in Table~\ref{tab:learning-comparison} in the appendix, highlighting their respective advantages, limitations, and use cases.}This allows LLMs to be applied seamlessly across diverse classification tasks (\cite{kaplan2020scaling}). While ICL offers significant advantages, its effectiveness heavily depends on handcrafted prompts which introduce several challenges. Manually designed prompts through prompt engineering remain an ad-hoc and trial-and-error process, requiring domain expertise to craft effective inputs. Despite its accessibility, prompt engineering exhibits black-box characteristics, as the exact way in which LLMs interpret and respond to prompts is often opaque. Small modifications in wording or structure can lead to unpredictable variations in output, making it difficult to establish clear causal relationships. Additionally, the effectiveness of a prompt is often model-specific, meaning that a well-performing prompt for one LLM may not generalize across different architectures. These limitations underscore the need for systematic and automated prompt optimization to improve consistency and performance.

Our proposed three-stage LLM-based framework addresses key limitations of existing text classification approaches by introducing automatic prompt optimization, dynamic exemplar selection, and a consensus mechanism. These methods, which have been adopted and adapted from computer science, are systematically integrated into political science research to enhance LLM performance. Our contributions include introducing these concepts to political science, integrating them into a unified framework, providing a simple and accessible implementation, and streamlining techniques by excluding unnecessary complexity for this task. This framework directly addresses critical challenges in LLM-based text classification. Automatic prompt optimization overcomes the trial-and-error nature of prompt engineering by systematically refining prompts, reducing reliance on manual tuning, and improving consistency. Dynamic exemplar selection mitigates the issue of LLM sensitivity to few-shot examples\footnote{Figure~\ref{fig:ZeroFewShotComparison} presents a comparison of zero-shot and few-shot prompting, demonstrating how structured inputs guide LLM decision-making.} by ensuring that the most relevant examples are selected for each classification task, leading to more stable and reliable outputs. The consensus mechanism improves classification reliability by reducing variability in LLM-generated responses, addressing inconsistencies, and improving agreement across model outputs.\footnote{A summary of how our framework differs from current practices in political science is provided in Table~\ref{tab:our contribution} in the appendix.}

Existing prompt engineering relies on intuition-driven trial and error, offering little feedback on where and how prompts fail, making generalization difficult. To address this, we propose using \textbf{automatic prompt optimization} in the first stage, which systematically refines prompts using human-labeled exemplars rather than subjective adjustments. Our data-driven approach curates a small set of labeled examples (typically <100) to generate an optimized prompt tailored to task requirements. This method minimizes guesswork, enhances replicability, and improves classification accuracy over prompt engineering. The effectiveness of our method is demonstrated through quantitative and qualitative evaluations.

The second stage, \textbf{dynamic exemplar selection}, tailors the prompt exemplars to the most contextually relevant cases for each query rather than relying on a fixed set. This adaptation enhances classification accuracy by making the model’s inference process more context-sensitive. Importantly, the selected exemplars are not only the most relevant to the query text but also representative of the dataset as a whole, ensuring that the model learns from diverse and structurally significant examples. This is particularly crucial when dealing with complex data structures, where certain underrepresented patterns or relationships could otherwise be overlooked. By maintaining both query relevance and dataset representativeness, this method strengthens the model’s ability to handle subtle conceptual boundaries more effectively and generalize across diverse classification tasks.

In the final stage, our framework employs a \textbf{consensus mechanism} in which two LLMs independently generate predictions, and a third LLM serves as a \textbf{judge} to resolve any disagreements. Rather than simply aggregating outputs, the judge model evaluates the reasoning behind each prediction—often through a Chain-of-Thought (CoT) process—and determines the more accurate classification. This approach mirrors human annotation workflows, where a third reviewer adjudicates conflicts between two coders, thereby increasing reliability and interpretability. By incorporating a structured resolution step, the LLM-as-judge mechanism reduces classification noise and improves robustness, offering a scalable solution for tasks that require high precision and consistency.

 We apply our framework to three classification tasks: sentiment analysis, stance detection, and multi-category campaign ad tone classification. Our results show that automatic prompt optimization provides a clear alternative to trial-and-error prompt engineering, offering structured guidance to distinguish commonly conflated concepts—such as sentiment versus stance—without requiring new model training. In the tone classification task, our consensus mechanism further enhances reliability by mimicking multi-coder setups and flagging labeling inconsistencies, a key feature for ensuring data quality in subjective tasks. These applications demonstrate the versatility and robustness of our method for a wide range of political science classification problems.

\section{Literature Review}

\paragraph{Large Language Models for Text Classification.}

The evolution of BERT’s "pre-training and fine-tuning" paradigm led to the creation of Large Language Models (LLMs), a subset of Pre-trained Language Models (PLMs) characterized by an immense scale of parameters and training data. Following a principle known as the \textbf{scaling law} (\cite{kaplan2020scaling}), researchers observed that increasing both the size of the model (in terms of parameters) and the amount of training data leads to significant improvements in the model's overall performance across a wide variety of tasks. A key outcome of scaling is what are termed \textbf{emergent abilities}—complex skills such as interpreting complex instructions and performing multi-step reasoning (\cite{wei2022emergent}). Among these abilities is \textbf{in-context learning}, which allows LLMs to interpret and execute tasks based solely on prompt instructions, eliminating the need for task-specific re-training or fine-tuning (\cite{dong2022survey}). This capability enables LLMs like GPT-3 to leverage their pre-trained knowledge directly for new tasks, distinguishing them from earlier models that rely heavily on additional training to adapt to different applications.

 One subcategory of in-context learning is \textbf{zero-shot learning} (\cite{brown2020languagemodelsfewshotlearners,radford2019language}), where LLMs can perform tasks based purely on a description of the task, without requiring any labeled exemplars. For instance, GPT-3 can classify the sentiment of a sentence (positive or negative) based solely on a prompt describing the task. Additionally, LLMs support \textbf{few-shot learning} (\cite{brown2020languagemodelsfewshotlearners,min-etal-2022-rethinking}), where the inclusion of a handful of exemplars in the prompt enables the model to better understand and generalize the task, further improving accuracy. This flexibility makes LLMs especially valuable in fields like political science, where labeled data is often scarce. By generating text token by token, they maintain coherent context across long passages, enabling them to handle a range of tasks based on a single prompt. This capability provides political scientists with a powerful, adaptable tool for text analysis without the need for fine-tuning, positioning LLMs as a resource-efficient alternative to traditional methods (\cite{radford2019language}).

LLMs offer powerful capabilities for text classification, yet their effectiveness often depends heavily on well-designed prompts, a reliance that has led to the labor-intensive practice of prompt engineering (\cite{khattab2023dspy}). Research applying LLMs to social science tasks has underscored this challenge, showing that LLMs alone may not consistently achieve sufficient accuracy without tailored prompt adjustments or additional human oversight. For example, ChatGPT, when used to classify social media sentiment on HPV vaccination, exhibited higher accuracy for anti-vaccination messages but struggled with pro-vaccination sentiment, particularly for longer formats, indicating the model’s sensitivity to both content type and length (\cite{kim2024accuracy}). Similarly, efforts to classify policy documents into specific issue categories using GPT-3.5 and GPT-4 revealed that LLMs achieved only moderate accuracy without human intervention—ranging from 65–83\%—and required significant manual oversight for contentious or complex cases (\cite{gunes2023multiclass}). These studies suggest that while LLMs offer promising automation potential, optimal results often require extensive prompt customization and careful selection of exemplar data.

A key characteristic of in-context learning (ICL) is its ability to optimize the context provided before generation rather than modifying model parameters. Instead of requiring additional training, ICL conditions the model through carefully structured prompts that influence its responses. Studies have shown that well-structured few-shot prompts can significantly improve LLM performance in text classification (\cite{min-etal-2022-rethinking}), reinforcing the importance of effective context design in achieving reliable results.

The prompt engineering workflow (illustrated in Figure \ref{fig:prompt_procedure}) involves three main stages: creating an initial prompt, testing the prompt on the model to generate results, and then verifying those results extensively. If the output yields low accuracy, the user re-enters a \textbf{Revise and Try} loop, repeatedly adjusting the prompt without clear feedback on how each change impacts model behavior. This blind, trial-and-error process can be time-consuming and uncertain, as users often lack systematic guidance for prompt refinement. Once a satisfactory level of accuracy is achieved, the final prompt is established for consistent application.

Despite its utility, prompt engineering has notable limitations. Ensuring that a prompt is both clear and accurately interpreted by the model remains challenging, as users often rely on summary metrics like F1 scores to assess quality. However, these metrics offer little guidance on \textit{how} to improve prompts, leading to an iterative cycle of speculative revisions without assurance of improved performance. Moreover, prompt effectiveness tends to be model-specific; a prompt optimized for one LLM may perform poorly on another, limiting generalizability across different models and data contexts (\cite{khattab2023dspy}). Although in-context learning increases adaptability by using exemplars within the prompt, it often relies on static exemplars. As the model encounters varied or new text data, these static exemplars may not align well, potentially reducing output accuracy. 

Recent work has introduced automatic prompt optimization methods like DSPy (\cite{khattab2023dspy}), APE (\cite{zhou2022large}), OPRO (\cite{yang2024largelanguagemodelsoptimizers}), and EvoPrompt (\cite{guo2024optimizersbetteronellm}), which aim to improve prompt performance by searching for high-performing strings. For instance, DSPy organizes tasks into step-by-step pipelines using modules called "teleprompters" that fine-tune model behavior at each step. These multi-stage approaches work well for tasks requiring structured reasoning or complex retrieval, like multi-hop question answering, but are often computationally intensive and may not translate effectively to political science text classification. DSPy’s complex decomposition and reliance on advanced models make it resource-demanding, limiting its practicality for more context-sensitive text analysis, like assessing sentiment or stance in political texts, which often requires careful contextual interpretation and is common in smaller-scale research projects.

To meet the specific demands of political science text analysis, we introduce a novel approach incorporating \textbf{automatic prompt optimization}, \textbf{dynamic exemplar selection}, and a \textbf{consensus mechanism}, which we will explain in detail in the following methodology section.

\section{Methodology}

\subsection{Data Preprocessing}

Before applying automatic prompt optimization, we first process and structure the dataset to enable efficient exemplar selection. This preparation stage includes two key steps: converting texts into embeddings and reducing feature dimensionality. These transformations help organize the dataset to support dynamic exemplar selection while optimizing use of the LLM’s context window.

\paragraph{Converting Texts into Embeddings.}  
LLMs convert natural language into fixed-size embedding vectors, enabling comparison of semantic similarity. We apply this transformation to all unlabeled texts using models such as OpenAI's "text-embedding-3-small." Following (\cite{Steck_2024}), we use cosine distance to quantify similarity between embeddings:

\[
\text{cosine\_distance}(\mathbf{x}_i, \mathbf{x}_j) = 1 - \frac{\mathbf{x}_i \cdot \mathbf{x}_j}{\Vert\mathbf{x}_i\Vert \Vert\mathbf{x}_j\Vert}
\]

\paragraph{Feature Reduction with UMAP.}  
Uniform Manifold Approximation and Projection (UMAP) is a widely used dimensionality reduction technique that preserves relationships in high-dimensional data. Conceptually, it builds on traditions in political science like factor analysis (\cite{kim1978introduction}), multidimensional scaling (\cite{borg2007modern}), and correspondence analysis (\cite{greenacre2017correspondence}), which have been used to uncover latent dimensions in political behavior, such as ideological scaling from roll-call votes (\cite{poole1985spatial}) and survey data (\cite{aldrich1977method}). UMAP advances these techniques using non-linear methods, enabling detection of complex structure in text embeddings while remaining compatible with established scaling approaches.

In practice, UMAP is also widely supported in computer science, especially for high-dimensional data. We use UMAP (\cite{mcinnes2020umapuniformmanifoldapproximation}) with cosine distance to reduce text embeddings, improving efficiency in later stages. The reduced representations also enable Euclidean distance calculations, which are faster and more broadly supported in downstream applications. For further implementation details, see Appendix~\ref{appendix:method-details}.\footnote{While we use UMAP here, our method supports alternative techniques such as Principal Component Analysis (PCA) (\cite{wold1987principal}), t-SNE (\cite{maaten2008visualizing}), and Independent Component Analysis (ICA) (\cite{hyvarinen2000independent}), depending on dataset and task requirements.}

\subsection{Automatic Prompt Optimization}

After structuring the dataset through embedding generation and feature reduction, we apply automatic prompt optimization. LLMs face a significant constraint in their context window size, which limits the number of tokens \footnote{In a Large Language Model (LLM), a "token" is the smallest unit of text that the model processes, essentially representing a word or a sub-word, which is used as the building block for understanding and generating language} the model can process simultaneously in few-shot learning scenarios. Moreover, within this window, earlier tokens in a long sequence may receive less attention, further constraining the effective use of examples for inference. This limitation necessitates a method that not only selects exemplars that are highly relevant and representative of the broader dataset but also minimizes the number of tokens used. Such a method can mitigate the risk of earlier tokens being underweighted, thereby preserving their contribution to inference. To achieve this, we propose a novel approach: first, we select a pool of \(M\) representative examples for human labeling, where \(k < M \ll N\), with \(N\) representing the total number of examples under study and \(k\) denoting the number of exemplars included in the context window for few-shot in-context learning. Then, at inference time, when calling the commercial API to classify unlabeled query text, we dynamically select the most relevant \(k\) examples to augment the prompt. In the \textbf{automatic prompt optimization} stage, we systematically curate this \textbf{exemplar pool} for human labeling. This curated set serves two critical purposes: first, it provides the LLM with a structured reference to infer the underlying labeling rules, reducing ambiguity in classification. Second, it enables an adaptive selection mechanism, ensuring that the most contextually relevant examples are incorporated for each query, thereby enhancing classification accuracy and robustness. By creating this initial pool of human-annotated examples, we lay the foundation for more effective and efficient use of the LLM in subsequent stages, balancing the need for comprehensive knowledge injection with the constraints of the model's context window. In this stage, we prepare this exemplar pool, and here are the steps.

\begin{figure}[H]
\centering
\includegraphics[width=\textwidth]{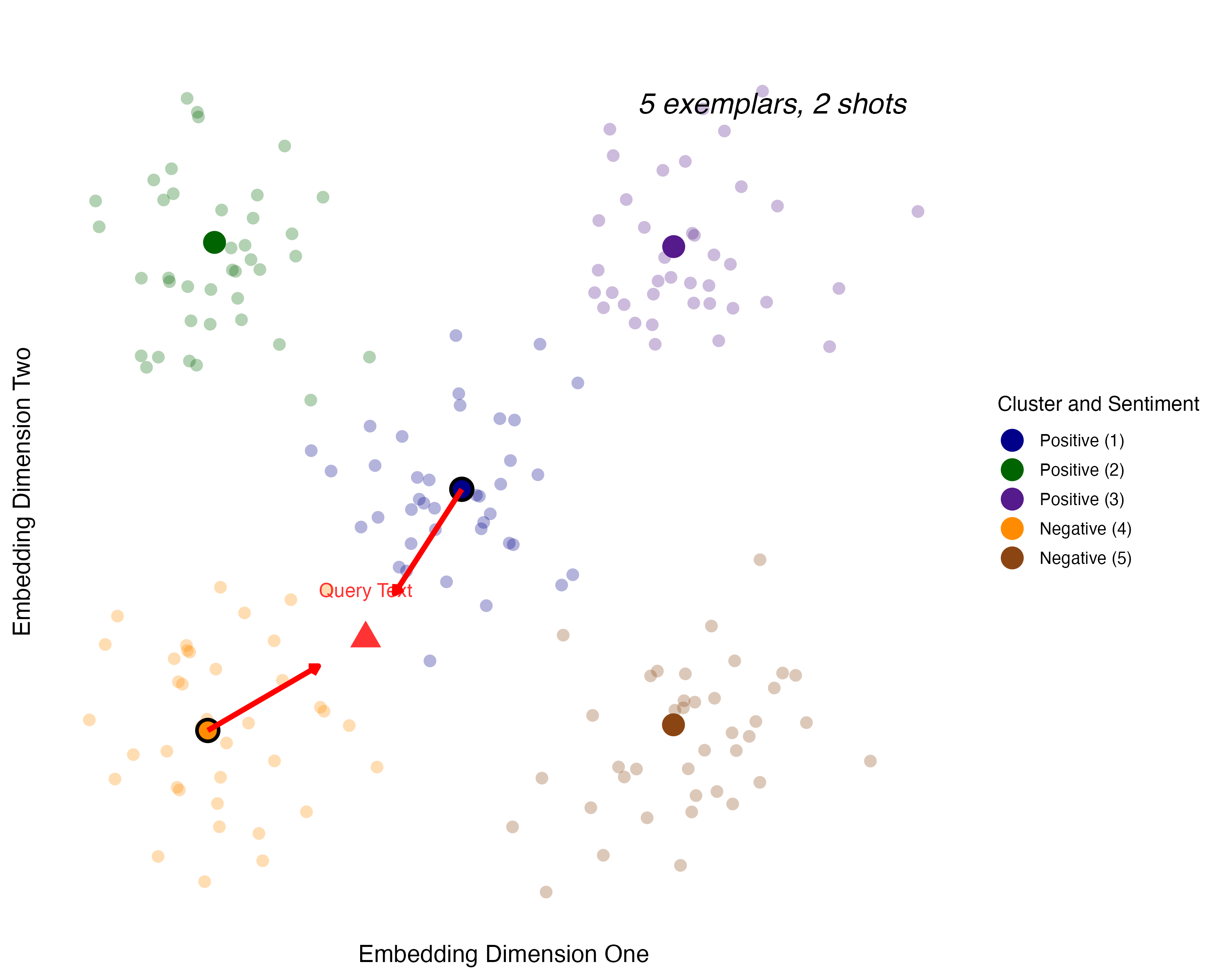}
\caption{Exemplar Selection Process for Few-Shot Learning: Demonstrating \textit{k}-means Clustering to Identify Representative Samples for Labeling}
\label{fig:exemplar_selection}
\end{figure}

\paragraph{Labeling Sample Selection.}  
To create a diverse and representative pool of examples, we employ an exemplar selection method (\cite{Bien_2011}) based on the manifold structure of the reduced embeddings. While various approaches exist for exemplar or prototype selection, such as set cover algorithms and density-based sampling, we opt for the \textit{k}-means selector due to its simplicity and effectiveness. This method involves performing \textit{k}-means clustering on the embeddings and then designating the text whose embedding is nearest to each cluster center as an exemplar. This approach ensures that every distinct group within the data is represented by one example, allowing for easy control over the number of exemplars selected while maintaining a comprehensive coverage of the embedding space. We keep \(M\) such texts as an exemplar pool and engage a human expert to label them. Typically, \(M < 100\), requiring minimal human effort. The process of labeling sample selection using \textit{k}-means clustering is illustrated in Figure~\ref{fig:exemplar_selection}. In this figure, we demonstrate a scenario where a pool of 5 exemplars is selected by \textit{k}-means clustering for labeling, indicating that \(M = 5\).\footnote{For further implementation details of the clustering and selection process, including the use of UMAP and \textit{k}-means, see Appendix~\ref{appendix:method-details}.}

\paragraph{Enhanced task description generation.} To enhance the initial task description, we leverage the LLM's analytical capabilities on the labeled exemplar pool. For each input-output pair from the pool, the LLM examines the rationale behind the human-assigned label. We then employ a Map-Reduce approach, where the LLM first "maps" by analyzing individual examples, and then "reduces" by summarizing the labeling rules for each class. This process ensures the generated rules are LLM-interpretable. Human experts can verify these rules for accuracy and intent. Typically, humans copy and append these generated rules to the initial prompt, creating an enhanced task description. In cases of inaccuracies, humans shall explicitly instruct the LLM to oppose specific incorrect rules in the prompt.

Our automatic prompt optimization method moves beyond prompt engineering by creating refined prompts based on human-labeled exemplars, rather than relying on intuition or trial and error.\footnote{Figure~\ref{fig:prompt_procedure} in the Appendix illustrates two workflows for finding optimal prompts in LLM applications: Prompt Engineering and our Automatic Prompt Optimization method. A detailed comparison between the two approaches is also available in Table~\ref{tab:prompt-comparison} in the Appendix, highlighting their respective processes.} Prompt engineering starts with an initial prompt, which is then tested in LLMs to generate results. These results are manually checked by researchers for consistency and accuracy. If accuracy is low, the researcher enters a "Revise and Try" loop, making adjustments based on intuition. This trial-and-error process is time-consuming and lacks clear feedback on where and how the prompt may have fallen short. As a result, creating a final satisfactory prompt often requires multiple speculative revisions, making it inefficient and hard to replicate. 

In contrast, our automatic prompt optimization method offers a systematic and transparent approach by using human-labeled exemplars rather than relying on intuition or trial and error. This process begins with researchers labeling a small set of curated examples, typically fewer than 100, which serve as a foundation for the model to generate an enhanced prompt tailored to the specific task criteria. This optimized prompt undergoes human verification, where it is assessed not only by numerical performance metrics, such as F1 scores, but also by its ability to accurately reflect human-labeled criteria and consistently produce reliable outputs. By grounding the prompt on well-defined exemplars, this workflow eliminates guesswork and minimizes trial and error. This structured, exemplar-based approach ensures effective and transparent prompt refinement.

\subsection{Dynamic Exemplar Selection}
In this stage, to address the limitations of using fixed exemplars in few-shot learning and their potential negative impact on classification outcomes, we propose a method to ensure that exemplars are both relevant to the specific query text and representative of the broader dataset. This dynamic selection process enhances the effectiveness of few-shot learning by tailoring exemplars to the specific input while maintaining comprehensive coverage of the dataset.

\paragraph{Few-shot example retrieval.} When calling LLM API for text classification, in a few-shot prompt setup, we retrieve an unlabeled query text and utilize its pre-computed embeddings to search for the top-k texts from a pool, selecting those with the highest scores using the Maximal Marginal Relevance (MMR) algorithm. The MMR (\cite{PARMAR2007879}) algorithm balances relevance and diversity by considering both the similarity between embeddings and the uniqueness of the selected examples, ensuring that the retrieved examples are not only relevant to the query but also varied enough to provide a comprehensive context.

Given a query embedding \( \mathbf{x}_q \), a pool \(R\) of exemplar texts, the MMR score for a candidate item \( \mathbf{x}_j \) from the pool \( R \) is defined as:

\[
\mathbf{x}_j = \argmax_{\mathbf{x}_j \in R \setminus S} \text{MMR}(R) \coloneqq \argmax_{\mathbf{x}_j \in R \setminus S} [ \lambda \cdot \text{Sim}(\mathbf{x}_q, \mathbf{x}_j) - (1 - \lambda) \cdot \max_{\mathbf{x}_i \in S} \text{Sim}(\mathbf{x}_j, \mathbf{x}_i) ],
\]

where \( S \) is the set of already selected items that is initially empty, and \( \lambda \) is a trade-off parameter between relevance and diversity (\(0 \leq \lambda \leq 1\)). In this way, we retrieve exemplars that are either semantically close to the query text with the same correct label or hard negatives that share some similarity with \( \mathbf{x}_q \) but from a different class. \ref{fig:exemplar_selection} illustrates a scenario where a query text is to be labeled, and the MMR (Maximal Marginal Relevance) algorithm selects two representative exemplars from the pool of five previously selected exemplars. This approach, known as two-shot learning, identifies the most relevant and similar exemplars to the query text, ensuring a contextually informed labeling process for few-shot learning.

\paragraph{Coarse Annotation with a Consensus Mechanism.}

Using a prompt enhanced with clearer task descriptions and carefully selected exemplars, we employ two LLMs to assign labels from predefined options. To improve accuracy and ensure reliability, we run the labeling process twice using two separate LLMs, allowing for comparison and verification of results. This approach reduces labeling inconsistencies by capturing high-confidence agreements while identifying hard cases where the models disagree. Discrepancies are tracked in a mismatch collection for further review, making the results more robust and interpretable.

\subsection{Consensus Mechanism}

In this stage, we utilize LLMs with in-context learning techniques to address mismatches identified in the dynamic exemplar selection stage. The prompts used here are refined versions of those from the previous stage, incorporating enhancements for fine-grained annotation and versatility. While these prompts may be computationally intensive, they are applied only to a limited number of queries in the mismatch collection, ensuring cost efficiency.

\paragraph{Chain-of-Thought Prompting.} A chain-of-thought (CoT) prompt guides a large language model (LLM) through a step-by-step reasoning process to enhance its ability to tackle complex tasks (\cite{wei2023chainofthoughtpromptingelicitsreasoning}). To implement it, the LLM is instructed to first analyze the content according to the task description, providing reasoning at each step before delivering the final answer. This approach works by mimicking human problem-solving, breaking down tasks into smaller components, which helps the model grasp the underlying logic and produce more accurate responses by not only assigning a label but also offering the reasoning behind it. CoT prompting has emerged as one of the most successful methods for improving the reasoning and inferential abilities of LLMs, making it a cornerstone of advanced prompt engineering (\cite{kojima2022large}). The output of CoT is a sequence \( y = (y_1, y_2, \dots, y_t) \) tokens, where \(y_t\) is the desired prediction and \(y_{<t}\) are the reasons:

\[
P(y_t | y_{<t}, \mathbf{x}, \text{prompt}; \theta)
\]

\paragraph{LLM-Based Consensus Arbitration.}
An LLM can serve as a judge (\cite{zheng2023judgingllmasajudgemtbenchchatbot}), leveraging its reasoning capabilities to evaluate and validate the outputs of other models. In our approach, two LLMs first annotate and reason through a small mismatch set using a Chain-of-Thought (CoT) process. The judging LLM then analyzes the reasoning steps provided by both models, assessing the quality, accuracy, and consistency of their responses. By evaluating the justification behind each prediction, rather than just the final label, this process enhances interpretability and ensures a more rigorous validation. This arbitration mechanism mirrors a multi-coder scenario, where two coders may disagree, and a third adjudicates discrepancies to ensure classification reliability. Previous research has demonstrated that incorporating LLMs as evaluators improves performance in high-precision tasks requiring structured reasoning and interpretability (\cite{zheng2023judgingllmasajudgemtbenchchatbot}). By integrating CoT reasoning with systematic judgment, our approach enhances consistency, reduces ambiguity, and strengthens the robustness of model outputs.

\subsection{Summary of Framework}

\begin{figure}[H]
\centering
\includegraphics[width=\textwidth]{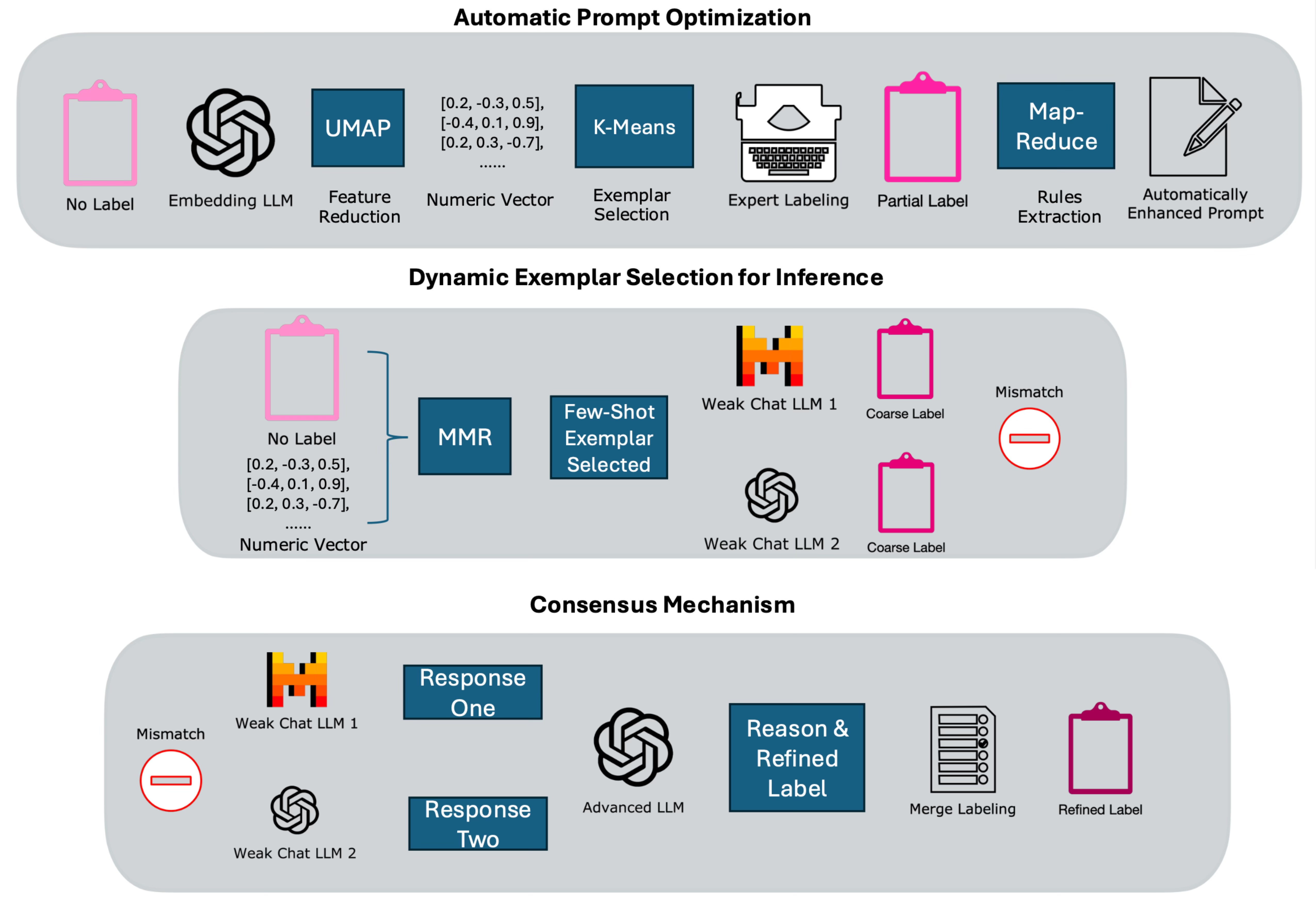}
\caption{Overview of the three core stages of our PoliPrompt framework}
\label{fig:three-stage_visualization}
\end{figure}

In this section, we summarize the entire framework, outlining each module's input-output and the human involvement required throughout the process as illustrated in Figure \ref{fig:three-stage_visualization}. The ``automatic prompt optimization'' stage prepares a pool of representative and diverse examples for human annotation in the subsequent stages. The input is human-collected, unlabeled texts, which we process using LLM embedding models, saving selected indices as an exemplar pool. Following this stage, human experts accurately annotate the texts in the pool and draft an initial prompt that includes only a task description. The ``dynamic exemplar selection'' stage focuses on enhancing this initial prompt with a refined task description and more appropriate examples. Here, two LLMs generate coarse annotations for the unlabeled examples. Users are responsible for running the task description generator, verifying its validity, and appending it to the initial prompt. After labeling, humans clean the predictions, identify mismatches, and record their indices in the dataset. The consensus mechanism then refines the predictions for these mismatches and provides reasoning for the adjustments. Humans review and clean the responses, replacing coarse predictions with fine-grained labels or conducting manual evaluations supported by LLM-generated reasoning.

By following this framework, we first leverage the in-context learning capabilities of large language models (LLMs), eliminating the need for intensive feature engineering, large labeled training datasets, task-specific training or fine-tuning, and high computational resources like GPUs. Therefore, this method addresses the limitations of commonly used text classification methods in political science. Second, we overcome two key limitations of LLMs: their heavy reliance on well-crafted prompts and the static nature of exemplars in few-shot learning. Our method dynamically optimizes prompts and selects relevant exemplars, making LLMs a more powerful and reliable tool for text classification in political science research.

\section{Experiments}
\paragraph{Overview.} We evaluated our method across three classification tasks in political science: sentiment analysis and stance detection on Twitter data related to Brett Kavanaugh's Supreme Court confirmation, and multi-category tone classification for campaign ads from the 2018 U.S. elections. The Twitter experiments demonstrate that our method can reliably distinguish between closely related concepts such as sentiment and stance—an area where many classifiers traditionally struggle. Furthermore, the campaign ad experiments highlight how the consensus mechanism enhances label consistency and helps detect potential noise in human annotations, particularly in more subjective or complex multi-category settings.

For sentiment and stance detection, Twitter was chosen for several key reasons. First, as one of the most widely used social media platforms, Twitter provides a wealth of real-time information during major events. For example, it has been a primary source of breaking news, especially during recent U.S. Presidential Elections (\cite{allcott2017social}). The unstructured data on Twitter offers valuable insights into public opinion and attitudes toward political figures and events (\cite{evans2014twitter}). Second, sentiment analysis and stance detection are crucial in political science for understanding public views on significant political topics or figures (\cite{hopkins2010method}, \cite{grimmer2013text}, \cite{gentzkow2019measuring}). Studies like those by \citet{hopkins2010method} and \citet{bond2015quantifying} show how analyzing sentiment and stance can reveal shifts in political attitudes and patterns of polarization. Finally, applying sentiment analysis and stance detection to the same dataset presents unique challenges, especially in distinguishing between the two. \citet{bestvater2023sentiment} argue that classifiers trained for sentiment analysis often struggle with stance classification, highlighting the challenge of distinguishing these frequently conflated political concepts. Our goal is to test whether our method can overcome this challenge, effectively separating sentiment from stance with high reliability. Specifically, we aim to achieve accurate stance detection through prompt optimization alone, without the need for retraining. 

We further applied our method to a multi-category classification task- analyzing campaign ads from the 2018 elections, categorizing them as contrast, attack, or promote. This task was chosen for several reasons. First, substantively, analyzing campaign ads provides valuable insights into campaign strategies and behavior across different media, including social media and traditional broadcast (\cite{fowler2021political}). Second, this task allows us to evaluate our method’s performance in multi-class classification, expanding beyond binary classification. Third, the longer text format of campaign ads compared to social media posts adds a layer of complexity, testing our method’s capacity to handle extended text\footnote{We conducted one last experiment involving multi-category topic classification on lengthy BBC news reports. This experiment demonstrates that our method is versatile and effective for classifying various types and lengths of text data, extending beyond short texts such as tweets. Detailed methodology and results are available in ~\ref{appendix:bbc-topic}.}
.

\paragraph{Setup.} Each dataset used in our experiments was manually labeled, serving as the ground-truth benchmark. In our experiments, we initially treated all texts as unlabeled, tasking the LLMs with predicting labels using our proposed method. We then compared the predicted labels to the human-labeled benchmark to evaluate the accuracy of our approach. For each dataset, we used the "text-embedding-3-small" model from OpenAI to convert all texts into embedding vectors. To reduce the dimensionality and improve efficiency, we applied UMAP to shrink each vector to 24 dimensions. We then selected 80 exemplars from each dataset using \textit{k}-Means clustering. These exemplars formed the example pool, which was leveraged to generate optimized prompts for LLM-based text classification\footnote{We conducted hyperparameter tuning experiments to assess the impact of varying the number of exemplars, testing sample sizes of 20, 40, 60, 80, and 100. Although F1 scores consistently improved as the number of exemplars increased, the rate of improvement slowed beyond a certain point. For our analysis, we chose to use only 80 exemplars in all experiments, deliberately demonstrating that our method can outperform traditional ML approaches even with a relatively modest number of examples. As a result, the performance metrics we report are conservative estimates, not reflecting the highest potential performance. Full results from the hyperparameter experiments can be found in Appendix 3.}. In our analysis, we define \textit{unoptimized application} as the direct application of LLMs without our optimization techniques and refer to their corresponding output as \textit{unoptimized results}. In contrast, results produced through the first two stages of our method—incorporating automatic prompt optimization and dynamic exemplar selection—are referred to as \textit{optimized results} for Mistral and GPT models. 

In our experiments on sentiment analysis and stance detection, we applied our method to classify over 3,660 tweets related to Brett Kavanaugh's Supreme Court confirmation, labeling each tweet as either "positive" or "negative" for sentiment and "support" or "oppose" for stance. We chose this task because prior research highlights the challenges of distinguishing sentiment from stance, as emphasized in the seminal article "Sentiment is Not Stance: Target-Aware Opinion Classification for Political Text Analysis" (\cite{bestvater2023sentiment}). By using automatic prompt optimization and supplying the model with 80 exemplars labeled for both sentiment and stance, we demonstrated that LLMs can effectively generate and refine clear definitions for these two dimensions. Additionally, the optimized prompt includes illustrative examples where sentiment and stance diverge, providing researchers with a reliable reference for crafting prompts that classify sentiment and stance independently. For researchers aware of the confusion surrounding concepts like sentiment and stance, our method shows that LLMs can effectively differentiate and summarize classification rules for these concepts when provided with a limited set of labeled exemplars. This demonstrates a significant advantage in transparency and explainability for researchers.

For sentiment classification, we implemented our three-stage method. Initially, the unoptimized LLMs, yielded low F1 scores. After applying automatic prompt optimization and dynamic exemplar selection, we observed a marked improvement, with F1 scores increasing by over 15\%. In the final stage, our consensus mechanism further enhanced accuracy, pushing F1 scores to nearly 92\%. These findings show that directly applying LLMs for political text classification often leads to suboptimal results, while each stage of our method delivers measurable gains in accuracy. Most importantly, the transparency offered by prompt optimization allows researchers to pinpoint possible sources of error in the prompts and revise them precisely, without relying on trial-and-error associated prompt engineering.

Following sentiment classification, we applied our method to stance detection, highlighting our approach’s effectiveness in classifying complex, often conflated concepts. Initially, the LLMs struggled to differentiate sentiment from stance, resulting in frequent misclassifications. Through automatic prompt optimization, however, the models dynamically generated an enhanced prompt that clarified the distinctions between "support" and "oppose," based on a set of 80 human-labeled tweets. This refinement yielded a dramatic improvement, with F1 scores increasing from 57\% to 95\%. 

In our final experiment, we attempted to replicate Fowler et al.'s (2021) study on how the medium of campaign ads influences tone. However, we encountered a highly noisy, human-labeled dataset, where labeling inconsistencies among human annotators became apparent. Our third stage of consensus mechanism effectively identified these conflicts, particularly in cases involving challenging or conceptually ambiguous classifications. This experiment highlights the significant impact a noisy dataset can have on downstream political analysis and demonstrates how our method can serve as a safeguard, flagging problematic labels and helping to prevent unreliable estimates early in the analysis process.

\subsection{Sentiment Analysis and Stance Detection toward Brett Kavanaugh's SCOTUS Nomination Using Twitter Data}

In their 2023 study, Bestvater and Monroe analyzed a hand-coded dataset of 3,660 tweets, labeling each for both sentiment (positive or negative emotional tone) and stance (support or oppose regarding an issue). To evaluate the ability of large language models (LLMs) to differentiate these two closely related dimensions, we selected a subset of 80 tweets from this dataset, containing \textbf{both} sentiment and stance labels, as input for automatic prompt optimization. This setup tested whether the LLM, guided by our approach, could generate an optimized prompt that effectively clarifies sentiment and stance.

Our initial prompt was straightward and unoptimized, as illustrated in Figure \ref{fig:prompt_sentiment_stance}. Through automatic prompt optimization, our method transformed this prompt into an enhanced version with guidelines, which not only provides definitions of sentiment and stance but also provides four clear examples illustrating different scenarios. This approach goes beyond basic instructions by integrating guidelines on special cases that commonly lead to confusion between sentiment and stance, such as instances where criticism of political opponents (rather than the primary subject, Brett Kavanaugh) could imply a negative sentiment without necessarily indicating opposition to Kavanaugh’s confirmation. By providing specific instructions for cases where sentiment and stance may diverge, this prompt allows the LLM to handle complex examples, such as sarcasm or irony, more effectively.

This prompt provides clear, interpretable distinctions between stance and sentiment, offering a reliable reference for researchers who may want to classify these concepts independently in future work. The prompt allows researchers to easily verify clarity in definitions, reducing the need for trial-and-error adjustments and enabling consistent, accurate classification.

\begin{figure}[H]
\centering
\caption{Comparison of Simple Heuristic and Enhanced Prompts for Sentiment and Stance Classification.}
\label{fig:prompt_sentiment_stance} 
\begin{tcolorbox}[
  colback=blue!5!white,
  colframe=blue!75!black,
  width=\textwidth,
  sharp corners=all,
  boxrule=0.8mm,
  title=Comparison of Simple Heuristic and Enhanced Prompts,
  fonttitle=\bfseries,
  colbacktitle=blue!10!white,
  coltitle=black
]

\begin{center}
    \textbf{\large Simple Heuristic Prompt}
\end{center}

Determine the \textbf{sentiment} and \textbf{stance} of the Twitter text related to Brett Kavanaugh’s Supreme Court confirmation.

Ensure your evaluation clearly differentiates between the \textbf{emotional sentiment} and the \textbf{political stance}.

\vspace{0.5cm} 

\noindent\color{black}\rule{\linewidth}{1pt} 

\begin{center} \textbf{\large Enhanced Prompt} \end{center}

You are given Twitter text about Brett Kavanaugh's confirmation. Your task is to determine both the \textbf{sentiment} (positive or negative) and the \textbf{stance} (support or oppose).

\textbf{Guidelines:} \begin{itemize} \item \textbf{Sentiment}: Assess the emotional tone, either \textcolor{purple}{\textbf{positive}} (e.g., \textit{\textcolor{purple}{approval}}) or \textcolor{purple}{\textbf{negative}} (e.g., \textit{\textcolor{purple}{condemnation}}), based on the language used. \item \textbf{Stance}: Identify the political position regarding Kavanaugh's confirmation, either \textcolor{teal}{\textbf{support}} (e.g., advocating for or defending the confirmation) or \textcolor{teal}{\textbf{oppose}} (e.g., arguing against or resisting the confirmation). \item \textbf{Note}: Sentiment and stance may not always align. For example, a \textcolor{purple}{positive emotional tone} may still reflect an \textcolor{teal}{opposing stance}, while a \textcolor{purple}{negative tone} could \textcolor{teal}{support} confirmation. \end{itemize}

\textbf{Key Indicators:} \begin{itemize} \item \textcolor{purple}{\textbf{Positive Sentiment}}: Words indicating \textit{\textcolor{purple}{happiness}} or \textit{\textcolor{purple}{affirmation}}. \item \textcolor{purple}{\textbf{Negative Sentiment}}: Words indicating \textit{\textcolor{purple}{condemnation}} or \textit{\textcolor{purple}{anger}}. \item \textcolor{teal}{\textbf{Support Stance}}: Text that defends, advocates for, or agrees with Kavanaugh’s confirmation. \item \textcolor{teal}{\textbf{Oppose Stance}}: Text that criticizes, argues against, or resists Kavanaugh’s confirmation. \end{itemize}

\textbf{Special Cases:} \begin{itemize} \item Humor or lighthearted criticism of the confirmation process itself, rather than the nominee, may indicate a \textcolor{purple}{\textit{positive sentiment}} but an \textcolor{teal}{\textit{opposing stance}}. \item Criticism of Kavanaugh's opponents or the process, rather than Kavanaugh himself, may suggest \textcolor{purple}{\textit{negative sentiment}} but a \textcolor{teal}{\textit{supportive stance}}. \end{itemize}

\textbf{Examples:} \begin{itemize} \item \textbf{Positive, Support}: "Kavanaugh is a great choice, he should be confirmed." \item \textbf{Positive, Oppose}: "Glad to see the protests against Kavanaugh." \item \textbf{Negative, Support}: "This investigation is unfair, but Kavanaugh will be confirmed." \item \textbf{Negative, Oppose}: "Kavanaugh's actions are disqualifying, he should not be confirmed." \end{itemize}

\end{tcolorbox}
\textit{Note: Red sections represent definitions and explanations for sentiment, while blue sections represent stance. This figure illustrates how, when prompted to treat sentiment and stance as distinct concepts, LLMs—guided by 80 carefully selected labeled exemplars—successfully differentiate between these two dimensions. The LLM-generated prompt provides detailed guidelines and examples, demonstrating a clear understanding of the unique characteristics of sentiment and stance.}
\end{figure}

In the next step, we used the same dataset as our ground-truth benchmark and tasked both GPT-3.5-turbo and Mistral-medium-latest with classifying \textbf{only} sentiment, starting with an unoptimized heuristic prompt based on the original instructions given to human coders by Bestvater and Monroe (\citealt{bestvater2023sentiment}). This unoptimized prompt is illustrated in Figure \ref{fig:Kavanaugh_sentiment_prompts}.

\begin{figure}[H]
\centering
\caption{Prompts for Sentiment Analysis about Brett Kavanaugh}
\begin{tcolorbox}[
  colback=blue!5!white,
  colframe=blue!75!black,
  title=Prompts for Sentiment Analysis towards Brett Kavanaugh,
  width=\textwidth,
  sharp corners=all,
  boxrule=0.8mm,
  fonttitle=\bfseries,
  colbacktitle=blue!10!white,
  coltitle=black
]

\begin{center}
\textbf{\large Unoptimized Heuristic Prompt:}\\
\end{center}

In your judgment, whether the specific sentiment is positive or negative? Please choose your answer only from the 2 options -- "positive" and "negative". Complete the task very succinctly using only one word written between '\textless' and '\textgreater'.\\

\noindent\color{black}\rule{\linewidth}{1pt}
\begin{center}
\textbf{\large Enhanced Prompt:}\\
\end{center}
In your judgment, whether the specific sentiment is positive or negative? Please choose your answer only from the 2 options -- "positive" and "negative". Complete the task very succinctly using only one word written between '\textless' and '\textgreater'. \textit{Evaluate the overall \textbf{emotional tone}, context, and specific language to determine if the sentiment is positive or negative.}\\

\vspace{0.3pt}

\textbf{\small negative:}\\
\textit{
Choose <negative> when the text expresses \textcolor{red}{\st{opposition}} \textcolor{blue}{\textbf{condemnation}}, criticism, dissatisfaction, frustration, anger, or uses negative language such as expletives, derogatory terms, or focuses on conflict, harm, or unethical behavior.}

\vspace{0.3pt}

\textbf{\small positive:}\\
\textit{
Choose <positive> when the text shows \textcolor{red}{\st{approval}} \textcolor{blue}{\textbf{affirmation}}, support, or uses favorable adjectives.}
\end{tcolorbox}
\label{fig:Kavanaugh_sentiment_prompts}
\vspace{0.5em} 

\textit{Note: Italicized sections were generated after applying automatic prompt optimization. The enhanced prompt succinctly summarizes the criteria for labeling text as "positive" or "negative," providing clarity and reducing ambiguity in sentiment classification.}
\end{figure}

\begin{figure}[H]
\centering
\includegraphics[width=\textwidth]{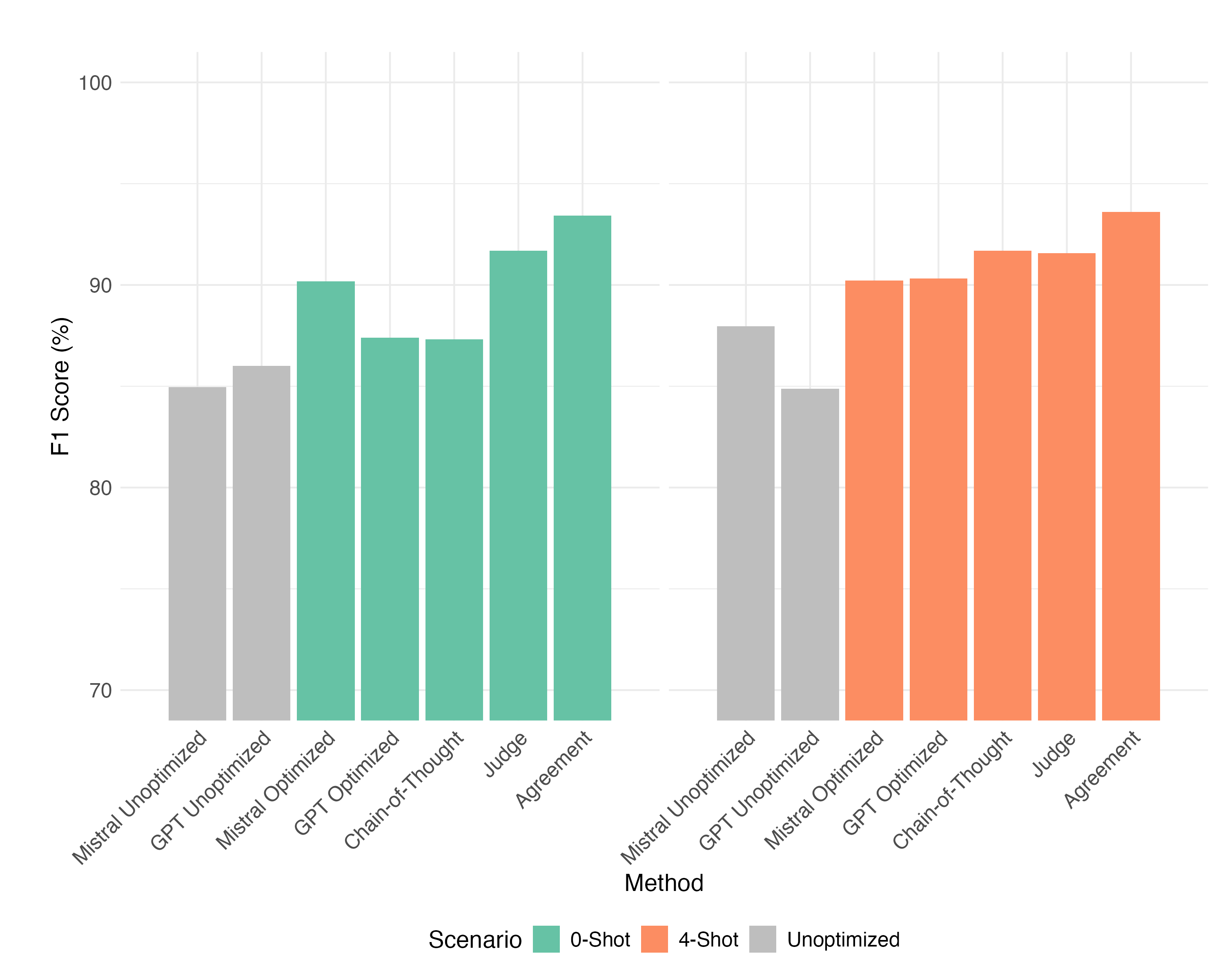}
\caption{Measuring Sentiment toward Kavanaugh: Comparison of F1 Scores across Different Methods}
\label{fig:f1_sentiment_Kavanaugh}
\begin{minipage}{\textwidth}
    \vspace{0.5cm} 
    \footnotesize
    \textit{\textbf{Note:} The gray bars represent the classification results using LLMs directly, without any enhancement from our method. The green and orange bars indicate the results after applying our three-stage method, demonstrating the effectiveness of each stage in improving classification performance.}
\end{minipage}
\end{figure}

Using this unoptimized prompt, GPT-3.5-turbo and Mistral-medium-latest achieved moderate F1 scores of 84\% and 88\%, respectively, as illustrated in Figure \ref{fig:f1_sentiment_Kavanaugh}. We then applied automatic prompt optimization. This enhanced prompt in Figure \ref{fig:Kavanaugh_sentiment_prompts} provided clearer criteria for identifying sentiment, focusing on emotional tone to improve classification precision. The optimized prompt increased GPT-3.5-turbo’s F1 score by over 5\% and Mistral’s by 3\%. When both models aligned in their predictions, the F1 score reached 94\%, as shown in the Agreement columns in Figure \ref{fig:f1_sentiment_Kavanaugh}. A third-stage consensus mechanism further refined classification reliability, highlighting the importance of structured prompt optimization and exemplar selection. This process illustrates the limitations of using unoptimized LLM outputs.

A notable outcome in applying our automatic prompt optimization method is the transparency it provides in distinguishing between stance and sentiment in prompts. Initially, the enhanced prompt generated for sentiment analysis included terms like “text expresses \textbf{opposition}, criticism…” for “negative” and “text shows \textbf{approval}, support, or uses favorable language” for “positive.” Even after using 80 representative, human-labeled exemplars, the LLMs continued to confuse stance with sentiment.

Thanks to the transparency of our method, we were able to directly examine the enhanced prompt, pinpoint the source of confusion, if any, and adjust the language accordingly. Specifically, we revised terms from “opposition” to “\textbf{condemnation}” and “approval” to “\textbf{affirmation},” guiding the model to focus on emotional tone rather than political stance. This refinement aligned the prompt with the definitions established in our previous application of automatic prompt optimization, as shown in Figure \ref{fig:prompt_sentiment_stance}, where positive sentiment encompasses happiness and affirmation, and negative sentiment includes terms indicating condemnation and anger. This example of sentiment classification highlights a clear advantage of our approach: by generating prompts grounded in structured, task-specific logic, our method allows researchers to systematically inspect and validate prompts without relying on the blind trial-and-error cycles typical of traditional prompt engineering. This transparency ensures that sentiment classification remains centered on emotional tone, enhancing interpretability, accuracy, and explainability by allowing researchers to understand what qualifies texts as positive or negative, including the specific emotions they express and the reasoning behind each classification.

Following sentiment classification, we extended our method to stance detection. Bestvater and Monroe’s study evaluated multiple text classifiers for stance detection and concluded that models or dictionaries initially designed for sentiment analysis often perform poorly on stance tasks, suggesting a need for models specifically trained for stance recognition. These findings are summarized in Table \ref{tab:metrics (Bestvater and Monroe)}. Contrary to Bestvater and Monroe’s recommendation that a newly trained model is essential for accurate stance classification, our method demonstrates that reliable classification can be achieved by simply changing and optimizing prompts.

As in the previous analysis, we began with the unoptimized heuristic prompt shown in Figure \ref{fig:Kavanaugh_prompts}. Using this unoptimized prompt, both Mistral-medium-latest and GPT-3.5-turbo achieved scores below 60\%, highlighting their difficulty in distinguishing between sentiment and stance. For example, both models incorrectly labeled the following tweet as “oppose”:

\begin{center} \textit{"RT @atensnut Democrats can’t just 'move on' and jump on the bandwagon of sketchy allegations against Kavanaugh without accepting the egregiousness of turning their backs on the victims of Bill Clinton."} \end{center}

The LLMs mistakenly categorized this as “oppose” because the tweet carries a tone of negativity and distrust, implicitly criticizing the Democratic Party for supporting the allegations against Kavanaugh while highlighting perceived hypocrisy. This example illustrates how LLMs, when prompted with a simple heuristic, can conflate sentiment with stance—misinterpreting negative sentiment toward a third party as opposition to the main subject. It underscores the importance of using structured, task-specific prompts that are designed to capture the nuanced distinctions required for accurate stance detection.

We then provided LLMs with \textbf{enhanced prompts} produced by our automatic prompt optimization method (Figure~\ref{fig:Kavanaugh_prompts}), offering clear and structured guidance that significantly improved their classification accuracy. Notably, the enhanced prompt goes beyond simple factual classification; for instance, it correctly identifies tweets as ``approve'' when they present evidence undermining Kavanaugh’s accusers. It also effectively captures and summarizes emotional cues that may imply support for Kavanaugh—such as \textbf{mocking} his opponents or expressing frustration at those obstructing his confirmation—nuances that are often misclassified as opposition when using heuristic-based prompting. These cases highlight how negative tone or sentiment alone can misleadingly suggest stance, underscoring the importance of prompts that distinguish between emotional expression and political position.

This refined prompt ensures both factual and emotional content are considered, leading to more accurate and nuanced classifications. We further validated the enhanced prompt against the definitions established in Figure~\ref{fig:prompt_sentiment_stance} and found strong alignment, reinforcing the conceptual distinction between sentiment and stance. The enhanced prompt yielded substantial performance improvements: Mistral-medium and GPT-3.5 achieved F1 scores of 91.69\% and 92.59\%, respectively—representing a \textasciitilde36\% increase over the initial zero-shot performance using a simple heuristic prompt, as shown in Figure~\ref{fig:f1_comparisons_Kavanaugh_stance}. In the four-shot\footnote{Detailed few-shot prompts are provided in Appendix~\ref{fig:Kavanaugh few-shot prompt}.} prompting scenario, the enhanced prompt enabled these models to achieve over a 20\% improvement, further demonstrating its value in producing accurate classifications.\footnote{The detailed resulting performance metrics for four-shot prompting are presented in Appendix~\ref{tab:Kavanaugh-4-shot-metrics}. Notably, by incorporating chain-of-thought reasoning and a final validation step using the Mistral-large-latest model as a judge, our approach surpassed traditional supervised classifiers, with F1 scores exceeding 95\%.}

\begin{figure}[H]
\centering
\caption{Zero-Shot Prompts for Analyzing Tweets about Brett Kavanaugh}
\begin{tcolorbox}[
  colback=blue!5!white,
  colframe=blue!75!black,
  title=Prompts for Analyzing Stance towards Kavanaugh,
  width=\textwidth,
  sharp corners=all,
  boxrule=0.8mm,
  fonttitle=\bfseries,
  colbacktitle=blue!10!white,
  coltitle=black
]

\begin{center}
\textbf{\large Simple Heuristic Prompt:}\\
\end{center}
In your judgment, whether the specific stance the author expresses toward the confirmation of Brett Kavanaugh is approving or opposing? Please choose your answer only from the 2 options -- "approve" and "oppose". Complete the task very succinctly using only one word written between '\textless' and '\textgreater'.\\

\noindent\color{black}\rule{\linewidth}{1pt}
\begin{center}
\textbf{\large Enhanced Prompt:}\\
\end{center}
\textit{You are a stance analyzer.} In your judgment, whether the specific stance the tweet text expresses toward the confirmation of Brett Kavanaugh is approving or opposing? \textit{Note: Focus on the stance expressed regarding Kavanaugh's confirmation. \textbf{Emotional tone} (e.g., anger, happiness) should be considered only if it directly influences the stance.} Please choose your answer only from the 2 options -- "approve" and "oppose". Complete the task very succinctly using only one word written between '\textless' and '\textgreater'.

\vspace{0.3pt}

\textbf{\small Oppose Stance:}\\
\textit{
- Lending credibility to allegations or accusations against Kavanaugh\\
- Highlighting potential disqualifying factors or controversies about Kavanaugh\\
- Expressing criticism, concerns, or questions about Kavanaugh's suitability\\
- Suggesting credible misconduct allegations should disqualify Kavanaugh as a nominee}

\vspace{0.3pt}

\textbf{\small Approve Stance:}\\
\textit{
- Discrediting or undermining accusations against Kavanaugh\\
- Presenting evidence that weakens the case against Kavanaugh\\
- Defending or rationalizing Kavanaugh’s nomination despite allegations\\
- Expressing frustration towards actions obstructing/delaying Kavanaugh's confirmation\\
- \textbf{Mocking}, dismissing, or discrediting Kavanaugh's accusers/opponents}
\end{tcolorbox}
\label{fig:Kavanaugh_prompts}
\vspace{0.5em} 

\textit{Note: Italicized sections reflect additions made after applying automatic prompt optimization. The optimized prompt succinctly outlines conditions under which text should be classified as “oppose” or “support.”}
\end{figure}

Our method demonstrates superior performance compared to unoptimized LLM predictions, largely due to the optimized prompt generated in the second stage of our approach. This shows an important consideration for political scientists: when using LLMs for text classification, relying on unoptimized labeling alone can lead to significant misclassifications, as LLMs may struggle with complex concepts like sentiment and stance. However, by providing a few exemplars and instructing the LLM to generate a more refined prompt with simple, targeted guidelines, we achieve outstanding results. Furthermore, our method outperforms both dictionary-based and traditional supervised learning approaches, all without the need to train a new model—contrary to the recommendation by Bestvater and Monroe (2023) for improved classification. We successfully minimized the cost associated with model training while maximizing classification accuracy.

\begin{figure}[H]
\centering
\includegraphics[width=\textwidth]{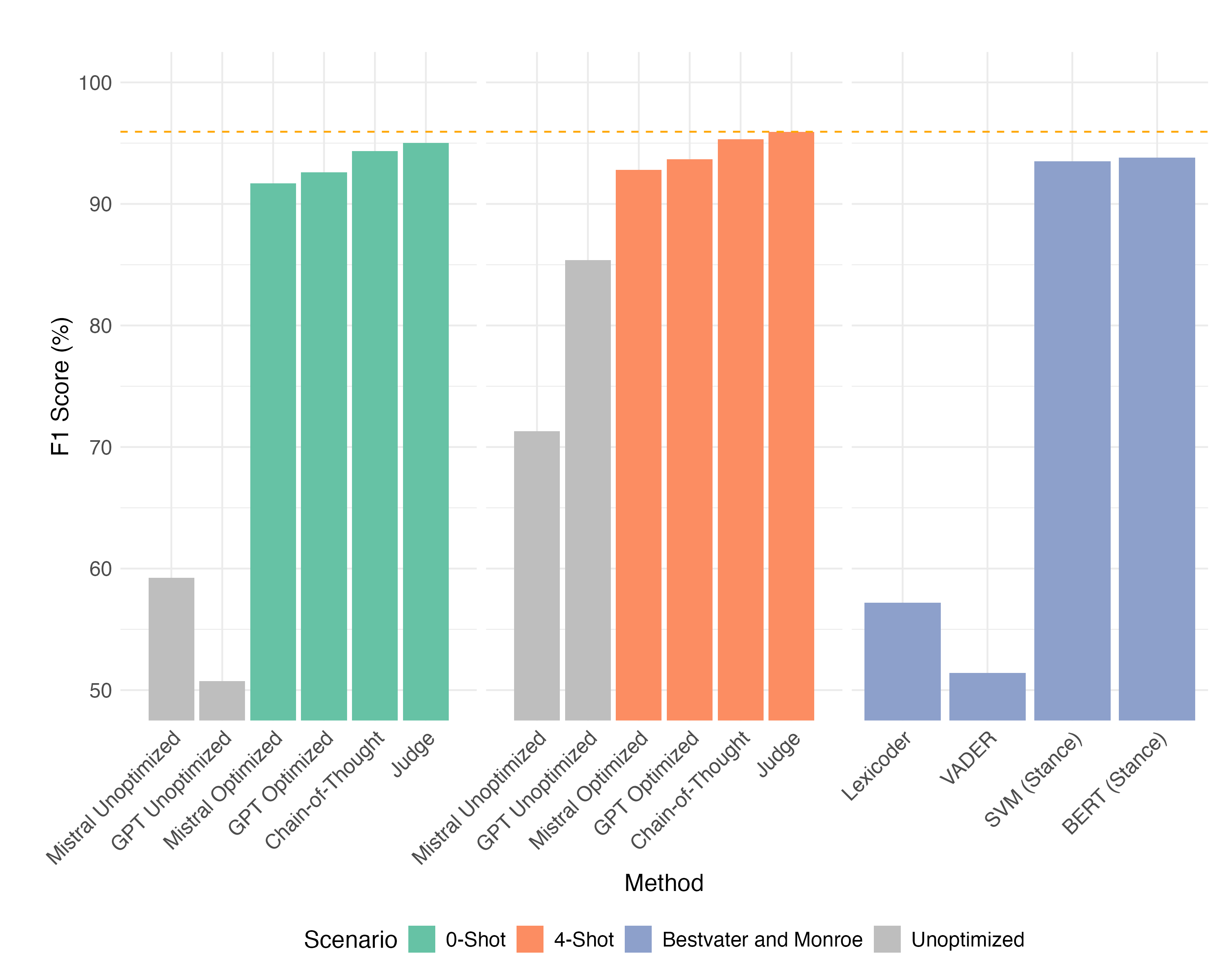}
\caption{Stance Detection toward Kavanaugh: Comparison of F1 Scores across Different Methods}
\label{fig:f1_comparisons_Kavanaugh_stance}

\end{figure}

We further assessed the relationship between sentiment and stance in our results by coding sentiment (positive as 1, negative as 0) and stance (support as 1, oppose as 0), using both \textbf{Pearson correlation} and the \textbf{Jaccard Index} to capture alignment between these two dimensions.\footnote{The Jaccard Index, which measures similarity by dividing the size of the intersection by the size of the union between sets, indicates greater overlap when higher, suggesting potential confusion in classifications between sentiment and stance (\cite{levandowsky1971distance}). Detailed formulas and interpretation for both metrics are provided in ~\ref{appendix:metrics-overall}.} For zero-shot results, the unoptimized application of LLMs showed a Pearson correlation of 0.38, whereas the optimized model returned a considerably lower correlation of 0.14, indicating weaker association between sentiment and stance. Similarly, the Jaccard Index for unoptimized zero-shot was 0.349, while optimization reduced it to 0.28, reinforcing that our method more effectively distinguishes stance from sentiment. A similar pattern emerged in the four-shot results: the unoptimized correlation was 0.31 compared to 0.07 in the optimized version, and the Jaccard Index dropped from 0.374 to 0.20. A statistical test confirmed that the difference in correlation coefficients was significant ($p < 0.05$), demonstrating that the optimized approach more effectively separates the two constructs.

\subsection{Classifying Campaign Ads Tones in the 2018 Election}

The medium through which political communication is delivered plays a critical role in shaping the message's tone and its audience reach. In a recent study, \citet{fowler2021political}  examined the impact of Facebook as a medium on the tone of political advertisements. They proposed that ads on Facebook are more likely to adopt a negative tone compared to other platforms\footnote{For a broader discussion on the prevalence and dynamics of negative campaign tone across various media, see Lau and Rovner's review article "Negative Campaigning" in the Annual Review of Political Science (\citeyear{lau2009negative}).}. 

To explore this hypothesis, the researchers collected data from political advertisements by all federal, statewide, and state legislative candidates during the 2018 elections. A team of research assistants then classified a sample of these ads based on their tone—whether they were \textit{promoting}, \textit{contrasting}, or \textit{attacking}. The dataset comprises a total of 14,642 advertisements, with 9,073 originating from Facebook and 5,569 from television ads, offering a comprehensive basis for comparing online and offline political messaging. We randomly selected a sample of 3,000 observations from the coded training set. Within this sample, 2,374 ads were classified as promoting a candidate, 448 as contrasting between candidates, and 178 as attacking a candidate\footnote{The original dataset exhibits a similar imbalance, with a significantly higher proportion of ads expressing a promotional tone. Importantly, our analysis shows that sampling 3,000 observations does not compromise the validity of downstream political analysis, as regression estimates derived from this subset are consistent with those reported in the original study.}.

We applied the same three-stage approach as in sentiment analysis and stance detection, this time focusing solely on reporting consensus mechanism results of the Chain-of-Thought (CoT) and Judge prompting methods. We evaluated their performance under both zero-shot and few-shot prompting conditions\footnote{Detailed unoptimized and optimized results are available in the Appendix.}. The final F1 scores, as shown in Table \ref{tab:summary_f1_scores_campaign_ads}, indicate notably low performance in the "attack" and "contrast" categories. Specifically, the average F1 score for "attack" is approximately 55\%, while for "contrast," it is even lower, averaging around 50\%. It is important to note that these F1 scores were calculated against gold-standard human labels, which are particularly noisy in this dataset, potentially contributing to the reduced accuracy.

\begin{table}[H]
\centering
\caption{Analyzing Ads Tones: Summary of F1 Scores across Different Methods}
\begin{tabular}{lcccc}
\toprule
\textbf{Class} & \textbf{0-Shot CoT} & \textbf{0-Shot Judge} & \textbf{6-Shot CoT} & \textbf{6-Shot Judge} \\
\midrule
\textbf{promote}  & 92.85\% & 93.23\% & 92.60\% & 93.14\% \\
\textbf{contrast} & 53.85\% & 45.67\% & 51.80\% & 49.78\% \\
\textbf{attack}   & 57.19\% & 53.68\% & 56.48\% & 58.56\% \\
\bottomrule
\end{tabular}
\begin{tablenotes}[hang]
\item[]\textit{Notes:} The reported figures represent the F1 scores across different categories, calculated after applying the third-stage chain-of-thought method and the judge model for labeling campaign advertisements. We noted that the F1 scores for both "contrast" and "attack" are particularly low.
\end{tablenotes}
\label{tab:summary_f1_scores_campaign_ads}
\end{table}

\begin{figure}[H]
\centering
\includegraphics[width=\textwidth]{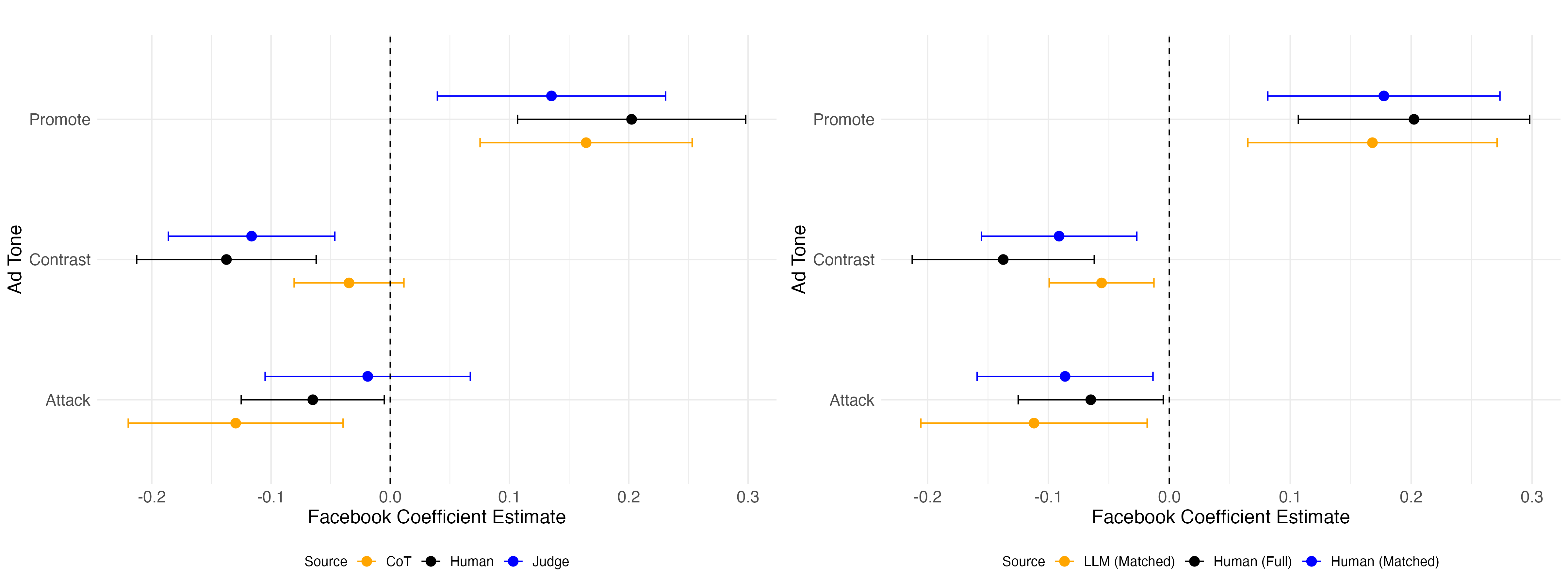}
\caption[Note]{Analyzing Ad Tones: Comparison of Estimates Using Different Labeling Approaches. \newline \textit{Notes:} The left panel of this figure shows the effect of medium on campaign ad tones across different labeling approaches, including initial human-labeled data before removing controversial labels}
\label{fig:downstream_comparisons_combined}
\end{figure}

Despite these lower F1 scores, we proceeded to use the predicted labels generated by zero-shot chain-of-thought prompting to re-estimate the same fixed-effect regression model described in Fowler's study. Specifically, we utilized a candidate-level fixed effects model, where the dependent variable is the average tone of the candidate's ads across various media platforms\footnote{Following Fowler et al. (2021), we computed expenditure-weighted averages of the message content for each candidate.}. Figure \ref{fig:downstream_comparisons_combined} illustrates the impact of different labeling strategies on downstream regression estimates. The left panel shows results obtained using zero-shot chain-of-thought prompting and the judge model, compared to fully human-labeled data. We observe that, except for the prompt tone, the estimates derived from LLM-labeled data diverge significantly from those obtained with human-labeled data. This divergence is expected, given that the F1 scores for these two categories are particularly low. 

These findings facilitated a re-examination of the data, particularly in cases where labels generated by multiple LLMs differ, mimicking a scenario with multiple human coders who occasionally disagree. Such discrepancies can highlight conceptual confusion or subtle ambiguities in the text, which may lead to controversial or inconsistent labeling decisions. Examples of such label inconsistencies, as shown in Table \ref{tab:label_discrepancies_main}, highlighting the need to question the consistency and reliability of the so-called gold standard human labels. In the first text, which was labeled as "contrast" by human coders, the primary focus is on criticizing career politicians for failing to prevent the destruction of industries, without directly contrasting specific policies with those proposed by others. On the contrary, the second piece, labeled as "promote" by human coders, explicitly criticizes the Republican Party in the state legislature while calling for support for other candidates. This message could reasonably be interpreted as both a promotion of alternatives and a contrast with the criticized party. Despite the similarities between these two ads—both of which criticize opponents and call for the support of others—the human labels differed. On the other hand, GPT and the final chain-of-thought method remained consistent in their classifications, demonstrating a more stable and coherent approach to labeling.

Given the noise within the human-labeled data, we proceeded to remove all controversial observations—those instances where two LLMs produced conflicting labels. Table~\ref{tab:f1_scores_comparison_adjusted} highlights the substantial improvements in F1 scores across all categories following this adjustment. Notably, the \textit{contrast} and \textit{attack} categories saw over a 30\% increase in F1 scores, demonstrating how filtering ambiguous cases can enhance label quality and model evaluation.

After this refinement, we reran the downstream fixed-effects model using outcomes exclusively from the optimized LLM classification on both matched data. The yellow line in the right panel of Figure \ref{fig:downstream_comparisons_combined} illustrates this estimator across each category. Additionally, we applied the fixed-effects model to this matched sample using human-labeled outcomes, shown by the blue lines in the same panel. The black line continues to represent estimates derived from human labels across the full sample. Notably, we observed that the estimates produced by LLM labeling on the matched sample closely align with those from human labeling on the matched sample. This similarity between LLM and human labeling on non-controversial data indicates the reliability of optimized LLM classifications in producing consistent results. Moreover, when comparing estimates derived from human labeling on the full sample of only 3,000 campaign ads to those based only on the matched sample, we observed notable differences, particularly for the Contrast category. This finding indicates that even within a full sample of 3,000 human-labeled ads, labeling inconsistencies can introduce noise that affects downstream analyses. Consequently, ensuring consistency in human labeling is crucial for achieving reliable estimates and minimizing potential biases in subsequent analysis.

Additionally, Table \ref{tab:summary_f1_scores_campaign_ads} revealed that zero-shot metrics surpassed few-shot performance. This anomaly can be attributed to the sampling of 80 examples for few-shot prompting during the dynamic exemplar selection stage. Given the noise within the human labels, these samples might have included incorrectly labeled examples, impacting the accuracy of few-shot promptings. This reliance on potentially flawed data could degrade the performance of few-shot prompts to levels below that of zero-shot prompts, which remain uninfluenced by human input. 

This experiment reveals a significant advantage of our method: its capacity to identify and flag inconsistencies in human labeling. Although human-coded data is often considered the “gold standard,” our findings reveal that this assumption can be problematic, particularly when dealing with contentious classifications. Human-coded classifications of ad tone are especially prone to noise due to the subjective nature of tone interpretation and the nuanced distinctions between categories like "attack" and "contrast." These categories often involve overlapping criteria, with small variations in content or emphasis leading to disagreements among coders. Additionally, individual biases, contextual interpretations, and a lack of universally agreed-upon definitions exacerbate inconsistencies in human annotations.\footnote{Inconsistencies among human coders are well-documented in the literature; see, for example, \citet{lombard2002content}, \citet{neuendorf2002content}, and \citet{artstein2008inter}. For ad tone classifications, the overlap between categories such as "attack" and "contrast" can further complicate consistent labeling, as highlighted in studies of political communication.}This analysis demonstrates that absolute reliance on human coding can be risky, especially when only a single coder is used. By detecting and flagging these discrepancies, our method enhances the robustness of classification results and allows researchers to address biases inherent in human-coded data before they affect downstream interpretations.

Furthermore, this experiment also reveals a potential pitfall of our approach. Specifically, when using our method—especially in few-shot prompting scenarios—researchers must ensure the accuracy of the exemplar text labeling. If the exemplars are noisy, it is highly likely to degrade the overall accuracy of the method.

\begin{table}[hbt!]
\centering
\caption{Analyzing Ads Tones: Examples of Inconsistencies and Controversies among Human and LLM Labels}
\begin{tabular}{p{4cm} p{0.8cm} p{0.8cm} p{0.8cm} p{5cm}}
\hline
\textbf{Text} & \textbf{Human} & \textbf{GPT} & \textbf{Mistral} & \textbf{GPT-4 CoT} \\ \hline
\textit{They wanted jobs - - - NOT HANDOUTS.  It was the Career Politicians job to prevent the destruction of an industry in his District. Vote ArtforCongress.com} & Contrast & Promote & Attack & Attack: The ad criticizes "Career Politicians" for failing to protect an industry, implying negligence or incompetence. It promotes "ArtforCongress.com" as an alternative, suggesting a vote for change. The focus is more on attacking the incumbents than contrasting policies or promoting Art extensively. \\ \hline
\textit{These are many of the common sense pieces of legislation that the GOP controlled Colorado Senate has killed, instead of working for Coloradans. This Tuesday, let’s elect folks who represent our Colorado values.} & Promote & Promote & Contrast & Attack: The ad criticizes the GOP controlled Colorado Senate for killing legislation, implying they do not represent Colorado values, and calls for electing representatives who do. It does not promote a specific candidate or contrast multiple candidates directly. \\ \hline
\end{tabular}
\label{tab:label_discrepancies_main}
\end{table}

\begin{table}[hbt!]
\centering
\caption{Analyzing Ads Tones: F1 Scores after Addressing Discrepancies: CoT Prompting vs. Judge Prompting (Zero-Shot)}
\begin{tabular}{lccc|ccc}
\toprule
\textbf{Class} & \multicolumn{3}{c|}{\textbf{CoT Prompting}} & \multicolumn{3}{c}{\textbf{Judge Prompting}} \\ \midrule
               & \textbf{Precision} & \textbf{Recall} & \textbf{F1 Score} & \textbf{Precision} & \textbf{Recall} & \textbf{F1 Score} \\ \midrule
\textbf{Promote}  & 0.9895 & 0.9520 & 0.9704 & 0.9838 & 0.9608 & 0.9721 \\ 
\textbf{Contrast} & 0.7163 & 0.7984 & 0.7551 & 0.7398 & 0.7280 & 0.7339 \\ 
\textbf{Attack}   & 0.7051 & 0.9322 & 0.8029 & 0.7094 & 0.9379 & 0.8078 
\\
\bottomrule
\end{tabular}
\begin{tablenotes}[hang]
\item[]\textit{Notes:} The reported figures represent the F1 scores across different categories, recalculated after removing the controversial human labels.
\end{tablenotes}
\label{tab:f1_scores_comparison_adjusted}
\end{table}

\section{Discussion and Conclusion}
In this paper, we introduce a novel three-stage LLM-based framework for text classification specifically tailored to political science research. Our approach addresses the challenges of traditional supervised learning and language models, which often require labor-intensive and computationally expensive feature engineering and fine-tuning. By leveraging inherent capabilities of in-context learning provided by LLMs, our method eliminates the need for such manual interventions. 

Crucially, our framework offers an innovative solution to one of the biggest challenges in using LLMs for text classification: the heavy reliance on prompt quality to achieve accurate outputs. Through \textbf{automatic prompt optimization}, our method allows LLMs to generate task-specific prompts that are precisely tailored to the task at hand, ensuring that models accurately interpret and classify the text. This optimization greatly improves the accuracy and reliability of classification results, minimizing the need for extensive human intervention in prompt design.

Additionally, our approach incorporates \textbf{dynamic exemplar selection}, which resolves the issue of irrelevant exemplar selection in few-shot learning. By dynamically choosing exemplars most relevant to the query text, our method prevents suboptimal classification results, further enhancing the overall performance and adaptability of the LLMs in diverse text classification tasks.

Finally, to ensure robustness and interpretability, our framework integrates a \textbf{consensus mechanism} in which multiple LLMs are used to cross-validate predictions. When discrepancies arise, a judging LLM evaluates the reasoning behind conflicting outputs to determine the most accurate classification. This structured adjudication process mirrors human multi-coder workflows and significantly reduces noise and uncertainty in model predictions, providing greater confidence in the final results.

Our method offers significant extensibility, making it a highly adaptable tool for political science research. One key advantage is its modular design, allowing researchers to easily customize or replace the algorithms within our framework to suit specific research needs. For example, although we utilized UMAP, \textit{k}-means, and MapReduce for exemplar selection and prompt generation in our experiments, these components are not fixed; they can be replaced with alternative algorithms tailored to the type of data or research objectives at hand. This flexibility makes the framework applicable to a wide variety of political science tasks and data types.

Moreover, our approach is designed to seamlessly integrate with the latest advancements in LLM technology, ensuring it remains cutting-edge without the need for significant modifications or added costs. For instance, if a newer model like GPT-x were to be released, researchers could easily incorporate it into our framework without any coding adjustments. This adaptability effectively future-proofs the method, enabling it to continually leverage state-of-the-art LLM capabilities. As long as LLMs are used, our method will remain applicable and relevant, providing a sustainable and up-to-date tool for political science research.

Last but not least, our method offers a cost-effective framework that achieves high accuracy through a consensus mechanism, making it a valuable tool for researchers seeking reliable yet accessible classification solutions.\footnote{Table~\ref{table:gpt_comparison} presents a cost and performance comparison between GPT-4-turbo and GPT-3.5-turbo. Our method, which relies on a consensus mechanism using GPT-3.5-turbo calls, achieves similar accuracy to GPT-4-turbo while reducing costs by more than a factor of 5 with three stages combined.}

\paragraph{Future Directions.}

While our method demonstrates strong performance in categorical text classification, it also opens up several avenues for future research. One promising direction involves extending the framework to tasks that require measuring intensity, degrees, or rankings—such as estimating political ideology on a continuous scale. For example, distinguishing between strong Democrats, moderate Democrats, Independents, Republicans, and strong Republicans on a 5-point Likert scale remains a challenge. This is due in part to the fact that large language models (LLMs) do not natively encode ordinal relationships during pre-training. Although they are highly effective at generating and classifying categorical language patterns, they lack a built-in understanding of ordinal or continuous gradations—a limitation that has long been recognized in the field of text scaling (see \cite{grimmer2013text}; \cite{lowe2013validating}). Developing methods that integrate ordinal awareness or continuous scaling into prompt-based classification could be especially impactful for political science applications.

Another important future direction is the extension of this framework beyond text to support multimodal data types such as images, videos, and audio, which are increasingly central to political communication and behavior research (e.g., \cite{Torres_Cantú_2022}; \cite{Girbau_Kobayashi_Renoust_Matsui_Satoh_2024}; \cite{Torres_2024}). Adapting the principles of automatic prompt optimization, dynamic exemplar selection, and LLM-based consensus to handle these diverse data forms would expand the utility of our approach and enable researchers to analyze a broader range of political content in real-world contexts.

\paragraph{Conclusion.}  
In conclusion, our proposed method leverages Large Language Models (LLMs) to enhance text classification in political science, addressing key limitations found in both traditional and LLM-based approaches. Unlike existing models that typically require task-specific retraining, our method achieves high accuracy through prompt adjustments alone. It uses \textbf{automatic prompt optimization} to produce task-specific prompts that clearly define complex political concepts, such as distinguishing sentiment from stance. This approach removes the guesswork typically associated with prompt engineering, providing researchers with a direct pathway to verify and refine prompts.

In addition, our \textbf{dynamic exemplar selection} method curates the most contextually relevant examples, enabling the model to adapt responsively to varying inputs. Together, the transparency and adaptability of our method ensure more reliable and interpretable classifications, streamlining the path from prompt design to application.

Moreover, our \textbf{consensus mechanism} combines the outputs of multiple LLMs to validate predictions and resolve conflicts. Rather than relying on a single model’s output, the method introduces a structured adjudication step that improves accuracy, interpretability, and robustness. This process is also valuable for detecting labeling inconsistencies—particularly in settings where only one human coder is available. By mimicking a multi-coder setup, our framework helps flag discrepancies early in the pipeline, ultimately enhancing both data quality and classification reliability. Finally, our method is implemented in an open-source Python package, \textbf{PoliPrompt}, ensuring accessibility and ease of use for researchers seeking to integrate these tools into their workflows.

\paragraph{Data Availability Statement}

Replication code and data for this article are available on GitHub. The materials will be uploaded to the Harvard Dataverse, and the corresponding citation and DOI will be provided in an updated version of this statement.

\paragraph{Competing Interests}

The authors declare none

\appendix
\counterwithin{figure}{section} 
\counterwithin{table}{section}  


\newcounter{Afigure}
\newcounter{Atable}
\newcounter{Bfigure}
\newcounter{Btable}
\newcounter{Cfigure}
\newcounter{Ctable}
\newcounter{Dfigure}
\newcounter{Dtable}

\renewcommand{\theAfigure}{A\arabic{Afigure}}
\renewcommand{\theAtable}{A\arabic{Atable}}
\renewcommand{\theBfigure}{B\arabic{Bfigure}}
\renewcommand{\theBtable}{B\arabic{Btable}}
\renewcommand{\theCfigure}{C\arabic{Cfigure}}
\renewcommand{\theCtable}{C\arabic{Ctable}}
\renewcommand{\theDfigure}{D\arabic{Dfigure}}
\renewcommand{\theDtable}{D\arabic{Dtable}}


\section{Resources for PoliPrompt Package Implementation}

\renewcommand{\thefigure}{\theAfigure} 
\renewcommand{\thetable}{\theAtable}   

This appendix provides direct links to the PoliPrompt package resources, including its GitHub repository, tutorial documentation, and PyPI listing for easy access and implementation.

\begin{itemize}
    \item \textbf{GitHub Repository:} \url{https://github.com/geshijoker/PoliPrompt} \\
    The GitHub repository includes the latest source code, detailed instructions for installation, and examples for using the package.

    \item \textbf{Tutorial Documentation:} \url{https://poliprompt-tutorial.readthedocs.io/en/latest/} \\
    Comprehensive tutorial documentation is available, guiding users through the implementation, setup, and features of PoliPrompt.

    \item \textbf{PyPI Page:} \url{https://pypi.org/project/PoliPrompt/} \\
    The package is available for installation via PyPI, making it straightforward to integrate PoliPrompt in Python projects.
\end{itemize}

These resources collectively provide a complete guide to installing, configuring, and utilizing PoliPrompt for research and analysis.

\section{Background on Large Language Models and Text Processing}

\renewcommand{\thefigure}{\theAfigure}
\renewcommand{\thetable}{\theAtable}

\begin{table}[]
\centering
\refstepcounter{Atable} 
\begin{tabular}{|p{2.5cm}|p{4cm}|p{4cm}|}
\hline
& \textbf{Current Common Practices in Political Science} & \textbf{Our framework} \\ \hline
\textbf{Task Description} & Manually Composed & Summarized by LLM from representative exemplars \\ \hline
\textbf{Few-shot examples} & Static examples across queries picked by human & Dynamically select examples for each query \\ \hline
\textbf{Solving Conflicts} & Simple average of predictions & Reaching consensus with an LLM as judge \\ \hline
\end{tabular}
\caption{The contributions of our framework}
\label{tab:our contribution}
\end{table}

\begin{table}[h]
\centering
\refstepcounter{Atable} 
\begin{tabular}{|p{2.5cm}|p{4cm}|p{4cm}|}
\hline
 & \textbf{In-Context Learning} & \textbf{Fine-Tuning} \\ \hline
\textbf{How it Works} & Uses prompts to guide the model & Modifies model parameters with training data \\ \hline
\textbf{Flexibility} & Adaptable to many tasks without retraining & Specializes the model for specific tasks \\ \hline
\textbf{Computational Requirement} & Doesn't require additional computational resources & Requires more computational power and data (e.g., GPUs) \\ \hline
\textbf{When to Use} & Good for prototyping and quick experimentation & Good for specialized domains and long-term use \\ \hline
\end{tabular}
\caption{Comparison between In-Context Learning and Fine-Tuning}
\label{tab:learning-comparison}
\end{table}

\begin{figure}[H]
\centering
\refstepcounter{Afigure} 
\caption{Comparisons of Zero-Shot and Few-Shot Prompting}
\includegraphics[width=\textwidth]{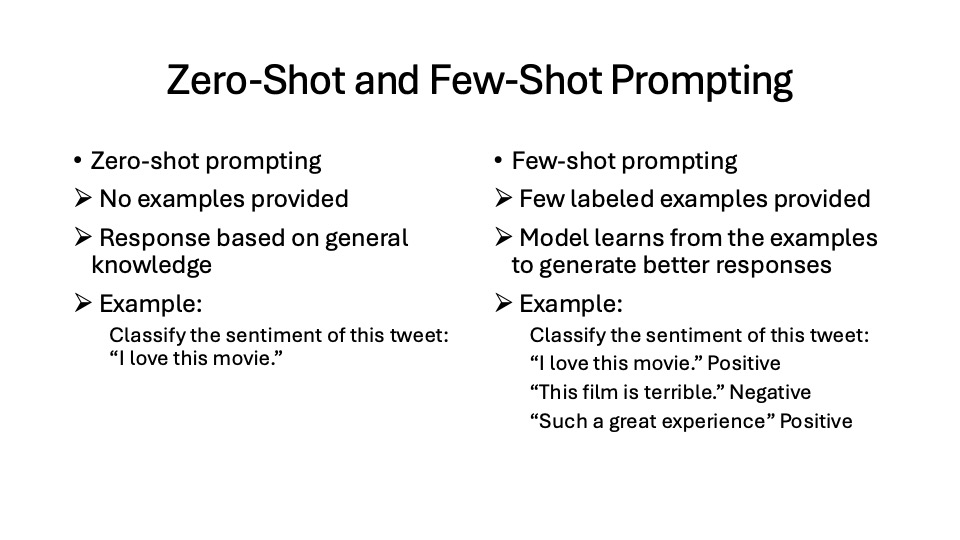}
\label{fig:ZeroFewShotComparison}
\end{figure}

\begin{table}[ht]
\centering
\refstepcounter{Atable} 
\begin{tabular}{|p{2cm}|p{4cm}|p{4cm}|}
\hline
\textbf{Aspect} & \textbf{Prompt Engineering} & \textbf{Automatic Prompt Optimization} \\ 
\hline
\textbf{Process} & Manual crafting by humans & Algorithmic/computational search \\ 
\hline
\textbf{Approach} & Trial and error, intuition-based & Systematic search using formal methods \\ 
\hline
\textbf{Based on} & Human expertise and domain knowledge & Quantitative metrics and objectives \\ 
\hline
\textbf{Time investment} & More time-intensive per prompt & Higher upfront compute cost, faster iteration \\ 
\hline
\textbf{Methods} & Templates, examples, specific instructions & Gradient-based optimization, RL, evolutionary algorithms \\ 
\hline
\textbf{Exploration scope} & Limited by human creativity & Can explore larger solution space \\ 
\hline
\end{tabular}
\caption{Comparison between Prompt Engineering and Automatic Prompt Optimization}
\label{tab:prompt-comparison}
\end{table}

\begin{figure}[htp]
\centering
\refstepcounter{Afigure}
\includegraphics[width=\textwidth]{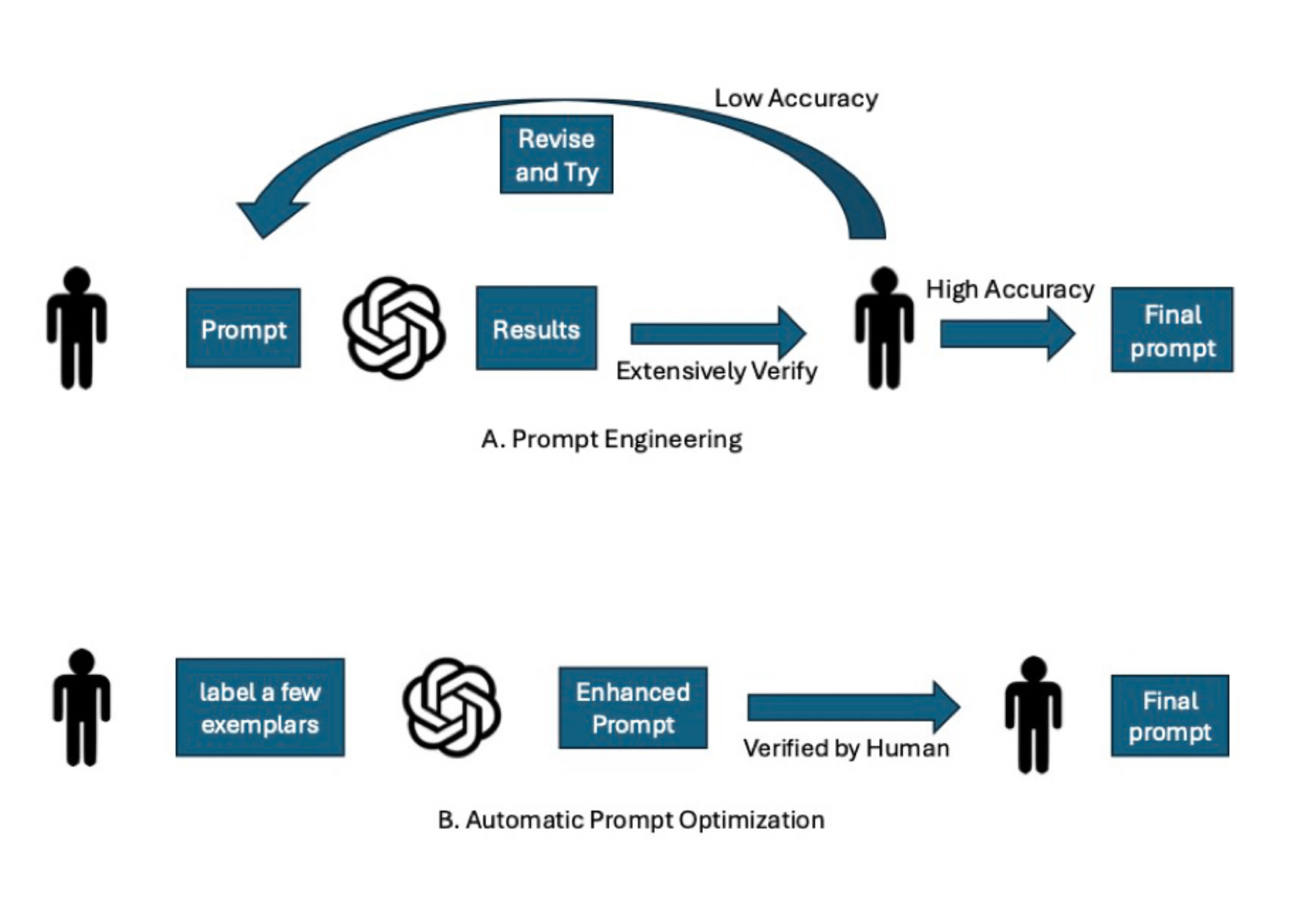}
\caption{The overview of Prompt Engineering and Automatic Prompt Optimization}
\label{fig:prompt_procedure}
\end{figure}

\begin{table}[hbt!]
\centering
\refstepcounter{Atable}
\begin{tabular}{lcc}
\toprule
\textbf{Feature} & \textbf{GPT-4-turbo} & \textbf{GPT-3.5-turbo} \\ 
\midrule
\textbf{Input Token Pricing} & \$10 / 1M tokens & \$0.5 / 1M tokens \\ 
\textbf{Output Token Pricing} & \$30 / 1M tokens & \$1.5 / 1M tokens \\ 
\textbf{Tokens per Minute} & 10,000 & 200,000 \\ 
\textbf{Batch Queue Limit} & 100,000 & 2,000,000 \\ \hline
\end{tabular}
\caption{Comparison of GPT-4-turbo and GPT-3.5-turbo at the time of writing (\cite{openai_pricing}).}
\label{table:gpt_comparison}
\end{table}

\section{Details of Methods}
\label{appendix:method-details}

This appendix provides additional details on the components used in our framework for exemplar selection and feature reduction.

\paragraph{UMAP with Cosine Distance.}  
To structure high-dimensional text embeddings, we apply Uniform Manifold Approximation and Projection (UMAP) for dimensionality reduction. UMAP preserves the semantic structure of the embedding space while reducing it to 2D or 3D (if specified) for clustering and visualization. We use cosine distance as the similarity metric during UMAP reduction, as it aligns better with how semantic similarity is distributed in text embeddings than Euclidean distance. Cosine distance is particularly suitable when working with normalized embedding vectors, capturing orientation rather than magnitude.

\paragraph{K-means Clustering with Euclidean Distance.}  
After dimensionality reduction, we apply K-means clustering to group embedding vectors into clusters of semantically similar examples. K-means minimizes within-cluster variance based on Euclidean distance, which is appropriate in the reduced space generated by UMAP. The resulting clusters guide the selection of a diverse yet representative pool of exemplars, ensuring better coverage of the data distribution.

\paragraph{Class-based Maximum Marginal Relevance (MMR).}  
To select the final set of $k$ exemplars for few-shot prompting, we implement a class-based Maximum Marginal Relevance (MMR) algorithm. MMR balances relevance to the current query (measured by cosine similarity in embedding space) with diversity among selected exemplars. We select examples that are both highly similar to the query and minimally redundant with one another. For each exemplar being selected, we prioritize the exemplars with the maximum MMR score but do not share the same class label with any selected exemplar in the set. This ensures that the prompt includes representative examples that capture intra-class variation while remaining relevant to the classification task.

\section{Metrics for Measurement}
\label{appendix:metrics-overall}

We use a variety of standard metrics to evaluate the performance and alignment of LLM-based classification tasks. Below are definitions and interpretations for each metric reported in the paper.

\paragraph{Accuracy.}  
Accuracy is the proportion of correct predictions among the total number of predictions:

\[
\text{Accuracy} = \frac{\text{Number of Correct Predictions}}{\text{Total Number of Predictions}}
\]

It provides a general sense of model performance, but may be misleading in imbalanced datasets.

\paragraph{Precision.}  
Precision measures how many of the predicted positive instances are actually correct:

\[
\text{Precision} = \frac{\text{True Positives}}{\text{True Positives + False Positives}}
\]

It reflects the model’s reliability when it predicts the positive class.

\paragraph{Recall.}  
Recall (or Sensitivity) measures how many of the actual positive instances the model correctly identifies:

\[
\text{Recall} = \frac{\text{True Positives}}{\text{True Positives + False Negatives}}
\]

It reflects the model’s ability to find all relevant cases in the dataset.

\paragraph{F1 Score.}  
The F1 Score is the harmonic mean of precision and recall, balancing the two:

\[
\text{F1 Score} = 2 \cdot \frac{\text{Precision} \cdot \text{Recall}}{\text{Precision + Recall}}
\]

It is especially useful for imbalanced classification tasks, where neither precision nor recall alone is sufficient.

\paragraph{Pearson Correlation.}  
The Pearson correlation coefficient $r$ measures the linear relationship between two variables. For binary classification tasks, it quantifies the extent to which two label sets co-vary:

\[
r = \frac{\sum_{i=1}^{n} (x_i - \bar{x})(y_i - \bar{y})}{\sqrt{\sum_{i=1}^{n}(x_i - \bar{x})^2} \cdot \sqrt{\sum_{i=1}^{n}(y_i - \bar{y})^2}}
\]

$r$ ranges from $-1$ (perfect negative correlation) to $1$ (perfect positive correlation), with $r = 0$ indicating no linear relationship.

\paragraph{Jaccard Index.}  
The Jaccard Index measures the similarity between two sets by comparing the size of their intersection over union:

\[
J(A, B) = \frac{|A \cap B|}{|A \cup B|}
\]

It ranges from 0 (no overlap) to 1 (perfect overlap). In binary classification, a lower Jaccard Index between different label types (e.g., sentiment vs. stance) suggests better conceptual separation.

\section{Topic Classification Experiment: Classifying BBC News Reports Topics}
\label{appendix:bbc-topic}

\renewcommand{\thefigure}{\theBfigure} 
\renewcommand{\thetable}{\theBtable}   

We also applied our method to a multi-category classification task involving extensive and lengthy text. We chose to label the topics of BBC news reports due to their diversity and relevance in benchmarking machine learning models. The dataset comprises 2,225 news articles sourced from the BBC News website, covering stories across five topical areas—business, entertainment, politics, sport, and tech—from the years 2004-2005. This dataset, originally compiled by Greene and Cunningham (2006), has been widely used in machine learning research as a benchmark for evaluating the performance of various classification algorithms. The diversity and structure of this dataset make it an ideal candidate for testing the robustness and accuracy of our proposed method in a real-world, multi-class classification scenario.

\begin{figure}[H]
\addtocounter{Bfigure}{1} 
\centering
\caption{Analyzing BBC Topics: Comparison of F1 Scores Across Different Methods}
\includegraphics[width=\textwidth]{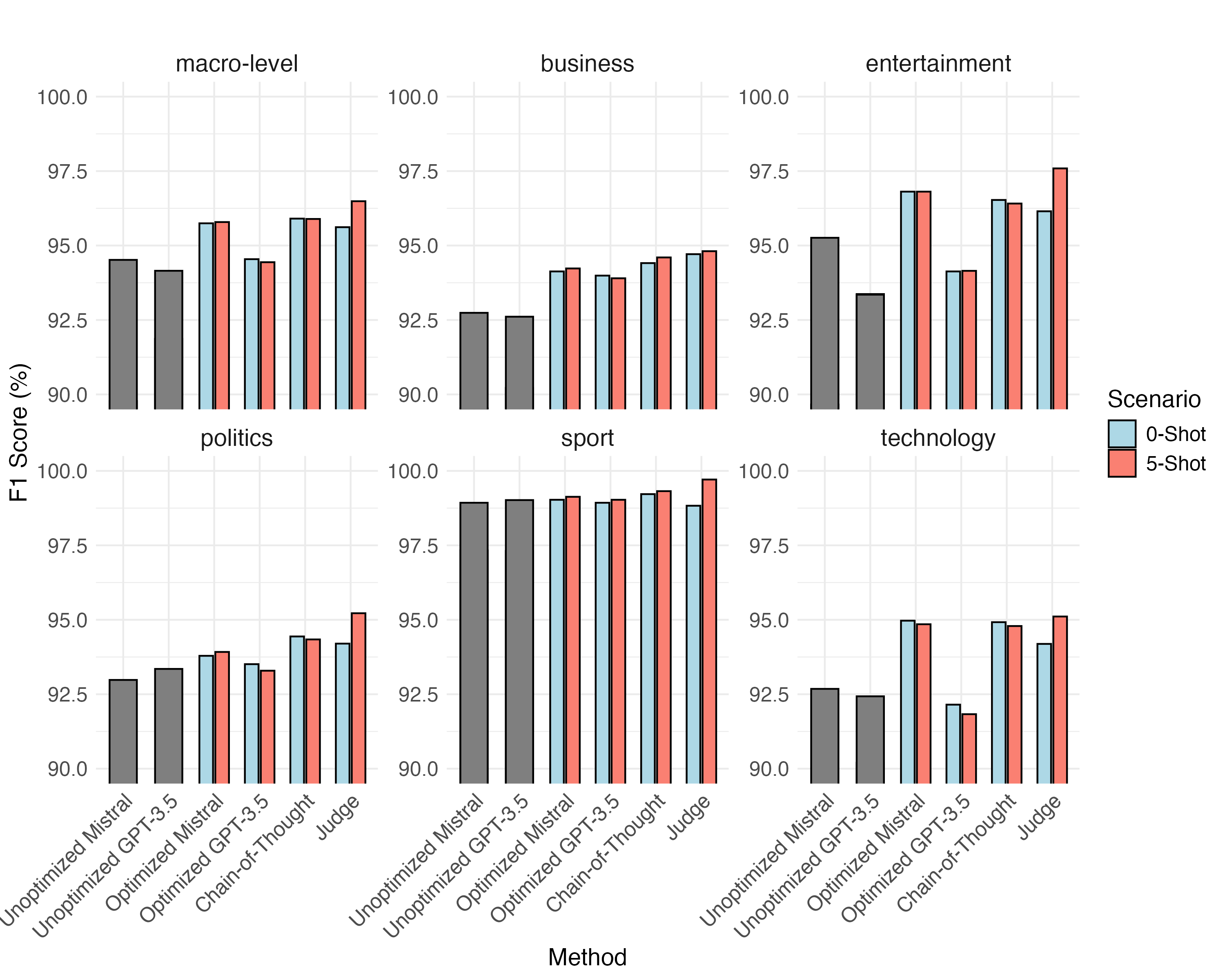}
\label{fig:f1_comparisons_topic}
\end{figure}

Starting with a simple heuristic prompt, as depicted in Figure \ref{fig:Topic_prompts}, we guided two weak LLMs to identify the primary topic of a news report from five categories: politics, business, sport, entertainment, and technology. Next, we employed a prompt generator, feeding the LLM with 80 exemplars labeled with accurate human classifications and asking the LLM to summarize the rules for categorizing news reports. This newly generated prompt was then fed back to the weak LLMs. As a result, the models not only corrected some of their initial misclassifications but also provided justifications for their decisions. In both zero-shot and five-shot settings\footnote{We employed five-shot prompting, providing the LLMs with one example per category since there are five topics to classify.}, these weaker models showed significant improvement in performance when we dynamically selected 80 exemplars compared to their initial unoptimized predictions\footnote{Detailed results are provided in Tables \ref{table:Topic_0shot} and \ref{table:Topic_5shot}.}. For example, as illustrated in Figure \ref{fig:f1_comparisons_topic} when using zero-shot prompting, both GPT-3.5 and Mistral-medium initially achieved F1 scores of approximately 0.89 in the "politics" category. However, these scores increased to around 0.94 after an enhanced prompt was generated. In categories like "sport," where the LLMs already performed exceptionally well with the naïve prompt—achieving F1 scores above 0.97—our method further boosted their performance. we were able to raise the F1 scores to around 0.99 putting into the enhanced prompt.

\begin{figure}[H]
\centering
\addtocounter{Bfigure}{1} 
\caption{An Example of Zero-Shot Prompts for Analyzing BBC News Reports}
\begin{tcolorbox}[colback=blue!5!white, colframe=blue!75!black, title=Prompts for Classifying Topics of BBC News Report, width=\textwidth]

\begin{center}
\textbf{\large Simple Heuristic Prompt:}\\  
\end{center}
In your judgment, what is the major topic of the given news report Text? Please choose only one answer from the following options: "politics", "business", "sport", "technology", or "entertainment". Complete the task very succinctly using only one word written between '\textless' and '\textgreater'.

\begin{center}
\textbf{\small [Enhanced Prompt]}\textbf{\small [BBC News Text (lengthy text in the Appendix~\ref{fig:Yahoo_News_Report})]}
\end{center}

\begin{tabularx}{\textwidth}{X|X} 
\textbf{Weak Chat Model1:} The news report discusses the history and future of Yahoo, including its competition with Google and MSN, its acquisitions, and its plans for the future. It also mentions the importance of search technology and the potential expansion into rich media. Based on this, the major topic of the news report is "technology." Therefore, the correct answer is <technology>.
& 
\textbf{Weak Chat Model2:} The text discusses the future prospects of Yahoo, a major internet company, focusing on its competition with Google and MSN, its business strategies, and its potential dominance in the next decade. Therefore, the correct answer is <business>.
\\
\end{tabularx}

\vspace{3pt}

\begin{center}
\textbf{\large Judge Prompt:}\\    
\end{center}
You are given 2 responses ("Response 1" and "Response 2") to the following task about the "Text", which can be correct or wrong. \\
Please judge which response (or neither) to the following content is correct step by step and provide your reasoning succinctly (Do not exceed 100 words). Finally, based on the reasoning, choose the correct answer and repeat the response's (or your) choice once in '\textless' and '\textgreater'.\\

\begin{center}
\textbf{\small [Response 1]}\textbf{\small [Response 2]}
\end{center}

\noindent\color{black}\rule{\linewidth}{1pt}
\textbf{Judge Chat Model:} Let's \textbf{think step-by-step}. Response 1 focuses on the technological aspects of Yahoo's history and future, such as search technology and expansion into rich media. However, the text primarily discusses Yahoo's business strategies, competition, and market position, rather than technological innovations. Response 2 correctly identifies the text's focus on Yahoo's business aspects, such as competition with Google and MSN, acquisitions, and future prospects. Therefore, the correct answer is <business>.\\
\end{tcolorbox}
\label{fig:Topic_prompts}
\end{figure}

In Figure \ref{fig:Topic_prompts}, we present an example where the two weak LLMs continued to differ in their classifications even after prompt enhancement. The full text of this news report example is available below in the \ref{fig:Yahoo_News_Report}. In this case, one model classified the news report as related to technology, while the other identified it as business. To resolve this discrepancy, we applied our third stage, which involved chain-of-thought prompting combined with a judge model. ``GPT-4-turbo'', as the judge model, ultimately classified the report as business, aligning with the human label. This decision was well-justified, as the judge model highlighted that the report, while contextualized within technology, primarily focused on business aspects.

The implementation of the third stage led to a notable improvement in performance metrics across all categories. For instance, when categorizing "politics," both chain-of-thought and judge prompting significantly boosted F1 scores in both zero-shot and five-shot scenarios, with an increase of approximately 0.1. Overall, the macro-level F1 score for all categories saw a substantial increase, reaching nearly 0.97.

In a dataset with accurate human labels, our method shows a progressive improvement in labeling accuracy, requiring only 80 labeled news reports to attain such high performance. This experiment underscores the efficiency and effectiveness of every stage in our approach.

\begin{figure}[H]
\addtocounter{Bfigure}{1}
\centering
\caption{Yahoo's Potential Dominance in the Next Decade}
\begin{tcolorbox}[
  colback=white,
  colframe=black,
  title=Appendix: News Report on Yahoo's Potential Dominance,
  width=\textwidth,
  sharp corners=all,
  boxrule=0.8mm,
  fonttitle=\bfseries,
  colbacktitle=gray!10!white,
  coltitle=black
]

\textbf{Can Yahoo dominate the next decade?}

Yahoo has reached the grand old age of 10 and, in internet years, that is a long time.

For many, Yahoo remains synonymous with the internet - a veteran that managed to ride the dot-com wave and the subsequent crash and maintain itself as one of the web's top brands. But for others there is another, newer net icon threatening to overshadow Yahoo in the post dot-com world - Google.

The veteran and the upstart have plenty in common - Yahoo was the first internet firm to offer initial public shares and Google was arguably the most watched IPO (Initial Public Offering) of the post-dot-com era. Both began life as search engines although in 2000, when Yahoo chose Google to power its search facility while it concentrated on its web portal business, it was very much Yahoo that commanded press attention. In recent years, the column inches have stacked up in Google's favour as the search engine also diversifies with the launch of services such as Gmail, its shopping channel Froogle and Google News.

For Jupiter analyst Olivier Beauvillain, Yahoo's initial decision to put its investment on search on hold was an error. "Yahoo was busy building a portal and while it was good to diversify they made a big mistake in outsourcing search to Google," he said "They thought Google would just be a technology provider but it has become a portal in its own right and a direct competitor," he added. He believes Yahoo failed to see how crucial search would become to internet users, something it has rediscovered in recent years. "It is interesting that in these last few years, it has refocused on search following the success of Google," he said. But for Allen Weiner, a research director at analyst firm Gartner and someone who has followed Yahoo's progress since the early years, the future of search is not going to be purely about the technology powering it. "Search technology is valuable but the next generation of search is going to be about premium content and the interface that users have to that content," he said. He believes the rivalry between Google and Yahoo is overblown and instead thinks the real battle is going to be between Yahoo and MSN. It is a battle that Yahoo is currently winning, he believes. "Microsoft has amazing assets including software capability and a global name but it has yet to show me it can create a rival product to Yahoo," he said.

He is convinced Yahoo remains the single most important brand on the world wide web.

"I believe Yahoo is the seminal brand on the web. If you are looking for a textbook definition of web portal then Yahoo is it," he said. It has achieved this dominance, Mr Weiner believes, by a canny combination of acquisitions such as that of Inktomi and Overture, and by avoiding direct involvement in either content creation or internet access. That is not to say that Yahoo hasn't had its dark days. When the dot-com bubble burst, it lost one-third of its revenue in a single year, bore a succession of losses and saw its market value fall from a peak of \$120bn to \$4.6bn at one point. Crucial to its survival was the decision to replace chief executive Tim Koogle with Terry Semel in May 2001, thinks Mr Weiner. His business savvy, coupled with the technical genius of founder Jerry Yang has proved a winning combination, he says.

So as the internet giant emerges from its first decade as a survivor, how will it fare as it enters its teenage years? "The game is theirs to lose and MSN is the only one that stands in the way of Yahoo's domination," predicted Mr Weiner. Nick Hazel, Yahoo's head of consumer services in the UK, thinks the fact that Yahoo has grown up with the first wave of the internet generation will stand it in good stead. Search will be a key focus as will making Yahoo Messenger available on mobiles, forging new broadband partnerships such as that with BT in the UK and continuing to provide a range of services beyond the desktop, he says. Mr Weiner thinks Yahoo's vision of becoming the ultimate gateway to the web will move increasing towards movies and television as more and more people get broadband access. "It will spread its portal wings to expand into rich media," he predicts.

\end{tcolorbox}
\label{fig:Yahoo_News_Report}
\end{figure}

\section{Hyperparamter Experiments on Analyzing BBC Topics}

\renewcommand{\thefigure}{\theCfigure} 
\renewcommand{\thetable}{\theCtable}   

\begin{table}[H]
\centering
\addtocounter{Ctable}{1}
\caption{Optimized Results of Mistral with 20 Exemplars}
\begin{tabular}{cccc}
\toprule
\textbf{Class} & \textbf{Precision} & \textbf{Recall} & \textbf{F1 Score} \\ \midrule
business & 0.9767 & 0.9039 & 0.9389 \\
entertainment & 0.9163 & 0.9922 & 0.9527 \\ 
politics & 0.9031 & 0.9832 & 0.9414 \\ 
sport & 0.9808 & 1.0000 & 0.9903 \\ 
technology & 0.9833 & 0.8828 & 0.9304 \\ 
\bottomrule
\end{tabular}
\label{tab:class_metrics}
\end{table}

\begin{table}[H]
\centering
\addtocounter{Ctable}{1}
\caption{Optimized Results of Mistral with 40 Exemplars}
\begin{tabular}{cccc}
\toprule
\textbf{Class} & \textbf{Precision} & \textbf{Recall} & \textbf{F1 Score} \\ \midrule
business & 0.9783 & 0.8843 & 0.9289 \\
entertainment & 0.9433 & 0.9922 & 0.9672 \\ 
politics & 0.9000 & 0.9712 & 0.9343 \\ 
sport & 0.9715 & 1.0000 & 0.9855 \\ 
technology & 0.9712 & 0.9252 & 0.9476 \\ 
\bottomrule
\end{tabular}
\label{tab:class_metrics_40_examples}
\end{table}

\begin{table}[H]
\centering
\addtocounter{Ctable}{1}
\caption{Optimized Results of Mistral with 60 Exemplars}
\begin{tabular}{cccc}
\toprule
\textbf{Class} & \textbf{Precision} & \textbf{Recall} & \textbf{F1 Score} \\ \midrule
\textbf{Business} & 0.9552 & 0.9197 & 0.9371 \\ 
\textbf{Entertainment} & 0.9793 & 0.9819 & 0.9806 \\ 
\textbf{Politics} & 0.9146 & 0.9760 & 0.9443 \\ 
\textbf{Sport} & 0.9751 & 0.9980 & 0.9865 \\ 
\textbf{Technology} & 0.9842 & 0.9302 & 0.9564 \\ 
\bottomrule
\end{tabular}
\label{tab:class_metrics_160}
\end{table}

\begin{table}[H]
\centering
\addtocounter{Ctable}{1}
\caption{Optimized Results of Mistral with 80 Exemplars}
\begin{tabular}{lccc}
\hline
\textbf{Class} & \textbf{Precision} & \textbf{Recall} & \textbf{F1 Score} \\
\hline
Business & 0.9646 & 0.9078 & 0.9354 \\
Entertainment & 0.9843 & 0.9741 & 0.9804 \\
Politics & 0.8965 & 0.9760 & 0.9358 \\
Sport & 0.9922 & 1.0000 & 0.9961 \\
Technology & 0.9772 & 0.9601 & 0.9686 \\
\hline
\end{tabular}
\label{tab:class_metrics_80}
\end{table}

\begin{table}[H]
\centering
\addtocounter{Ctable}{1}
\caption{Optimized Results of Mistral with 100 Exemplars}
\begin{tabular}{lccc}
\hline
\textbf{Class} & \textbf{Precision} & \textbf{Recall} & \textbf{F1 Score} \\
\hline
Business & 0.9713 & 0.9294 & 0.9499 \\
Entertainment & 0.9719 & 0.9845 & 0.9781 \\
Politics & 0.9292 & 0.9760 & 0.9520 \\
Sport & 0.9827 & 1.0000 & 0.9913 \\
Technology & 0.9794 & 0.9476 & 0.9632 \\
\hline
\end{tabular}
\label{tab:class_metrics_100}
\end{table}

\begin{figure}[H]
\centering
\addtocounter{Cfigure}{1}
\includegraphics[width=\textwidth]{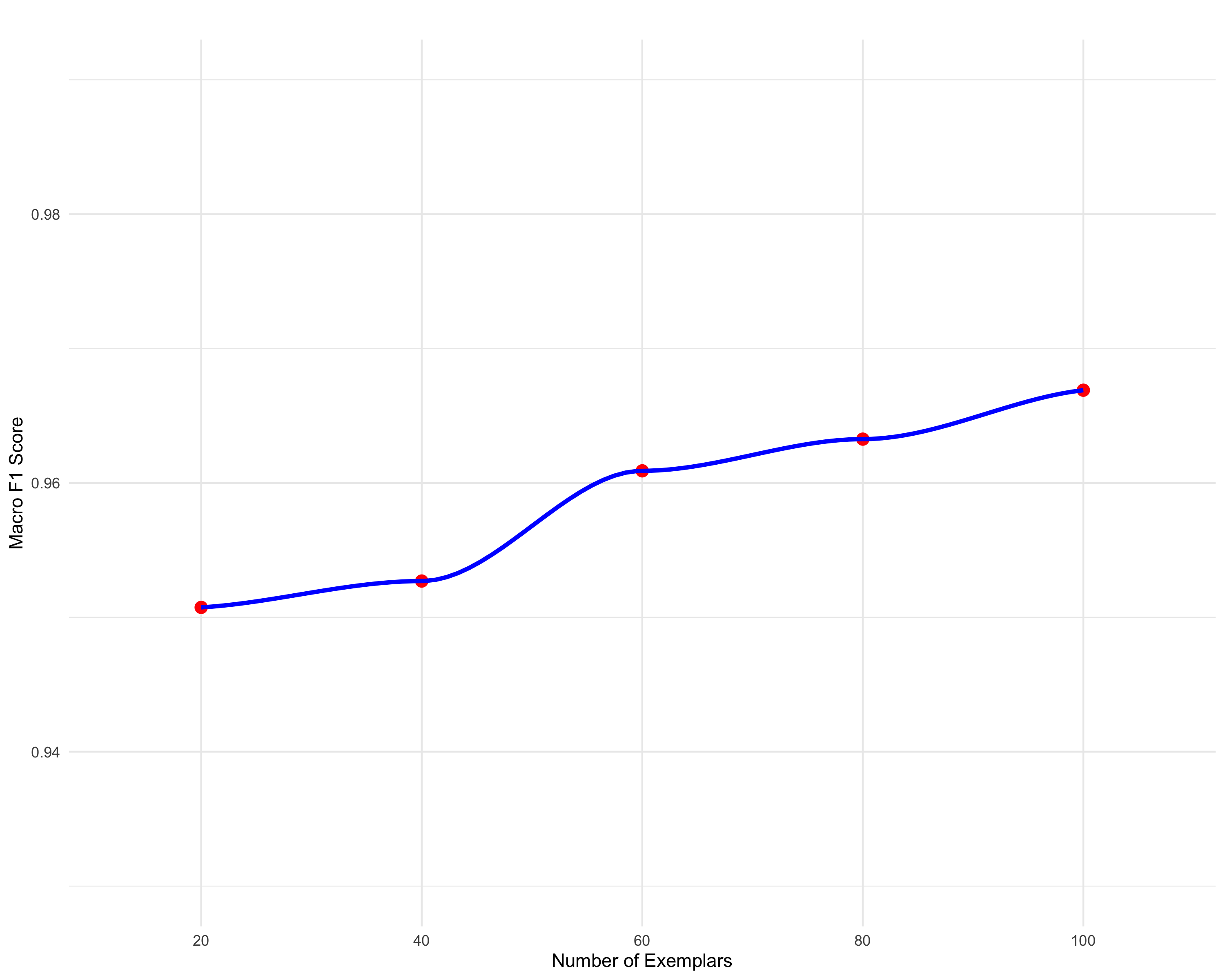}
\caption{BBC News Report Classification: F1 Scores of Different Numbers of Exemplars}
\label{fig:f1_comparisons_Kavanaugh}
\end{figure}

\clearpage

\section{Detailed Results across All Three Experiments}

\renewcommand{\thefigure}{\theDfigure} 
\renewcommand{\thetable}{\theDtable}   

\begin{table}[h]
\centering
\addtocounter{Dfigure}{1}
\caption{Performance Metrics in Bestvater and Monroe (2023)}
\begin{tabular}{lcc}
\toprule
\textbf{Approach} & \textbf{F1 Score (predicting stance)} \\ 
\midrule
Lexicoder & 0.572  \\ 
VADER & 0.514 \\ 
SVM (stance-trained) & 0.935  \\ 
BERT (stance-trained) & 0.938  \\
\bottomrule
\end{tabular}
\label{tab:metrics (Bestvater and Monroe)}
\end{table}

\begin{table}[H]
\centering
\addtocounter{Dtable}{1}
\caption{Brett Kavanaugh Experiment: Zero-Shot Performance Metrics}
\begin{tabular}{lccccc}
\toprule
\textbf{Approach} & \textbf{Accuracy (\%)} & \textbf{Precision (\%)} & \textbf{Recall (\%)} & \textbf{F1 Score (\%)} \\
\midrule 
Mistral (Unoptimized) & 0.6839 & 0.9882 &  0.4230 & 0.5925 \\ 
GPT-3.5 (Unoptimized) & 0.6333 & 0.9389 & 0.3476 & 0.5073 \\ 
Mistral (Optimized) & 0.9154 & 0.9801 & 0.8614 & 0.9169 \\ 
GPT-3.5 (Optimized) & 0.8887 & 0.9448 & 0.9078 & 0.9259 \\ 
CoT (153 Mismatches) & 0.9404 & 0.9717 & 0.9168 & 0.9434 \\ 
JUDGE (153 Mismatches) & 0.9473 & 0.9726 & 0.9291 & 0.9503 \\ 
\bottomrule
\end{tabular}
\label{tab:Kavanaugh 0-shot metrics}
\end{table}

\begin{figure}[H]
\centering
\addtocounter{Dfigure}{1}
\caption{Four-Shot Prompt for Analyzing Tweets about Brett Kavanaugh}  
\begin{tcolorbox}[colback=blue!5!white, colframe=blue!75!black, title=Prompts for Analyzing Stance towards Kavanaugh, width=\textwidth]

\textbf{\large Example 1:}\\
\textbf{Text:} \textit{RT @willchamberlain Ms. Ford sent an anonymous letter. She scrubbed her social media. She refuses to go on the record. She demanded an FBI investigation. She demands to testify after Kavanaugh. She demands no questions from outside counsel. Why? Because she’s lying. \#ConfirmKavanaugh}\\
\textbf{Answer:} approve

\vspace{0.2cm} 

\textbf{\large Example 2:}\\
\textbf{Text:} \textit{RT @johncardillo I just hope that when \#Kavanaugh is seated on SCOTUS, he remembers daily what Democrats and leftist activists did to him and his family.}\\
\textbf{Answer:} approve

\vspace{0.2cm} 

\textbf{\large Example 3:}\\
\textbf{Text:} \textit{RT @lanebrooks There seem to be many reasons for an FBI investigation into Kavanaugh. Sex assault. Concealed records. Mysterious payments. And we already know he lies under oath. Why is the GOP trying to ram this clown down our throats? https://t.co/BtQgRUWxWJ}\\
\textbf{Answer:} oppose

\vspace{0.2cm} 

\textbf{\large Example 4:}\\
\textbf{Text:} \textit{RT @PattyArquette I am going to remind you @SenatorCollins and all @GOP of what I said weeks ago. You are hanging your hat on Kavanaugh and you YOURSELVES haven’t read his records because they aren’t available. Is this the man you want to stake your political careers on?}\\
\textbf{Answer:} oppose

\end{tcolorbox}
\label{fig:Kavanaugh few-shot prompt}
\end{figure}

\begin{table}[H]
\centering
\addtocounter{Dtable}{1}
\caption{Brett Kavanaugh Experiment: Four-Shot Performance Metrics}
\begin{tabular}{lccccc}
\toprule
\textbf{Approach} & \textbf{Accuracy (\%)} & \textbf{Precision (\%)} & \textbf{Recall (\%)} & \textbf{F1 Score (\%)} \\ 
\midrule
Mistral (Unoptimized) & 0.75660  & 0.9875 &  0.5578 & 0.7130 \\ 
GPT-3.5 (Unoptimized) & 0.8530 & 0.9290 & 0.7897 & 0.8537 \\ 
Mistral (Optimized) & 0.9262 & 0.9864 & 0.8763 & 0.9281 \\ 
GPT-3.5 (Optimized) & 0.9331 & 0.9609 & 0.9140 & 0.9368 \\ 
CoT (271 Mismatches) & 0.9508 & 0.9808 & 0.9274 & 0.9533 \\ 
JUDGE (271 Mismatches) & 0.9568 & 0.9796 & 0.9401 & 0.9594 \\ 
\bottomrule
\end{tabular}
\label{tab:Kavanaugh-4-shot-metrics}
\end{table}

\begin{table}[H]
\centering
\addtocounter{Dtable}{1}
\caption{BBC Classification Experiment: F1 Scores for Zero-Shot Prompting}
\begin{tabular}{lccccc}
\toprule
\textbf{Approach} & \textbf{Business} & \textbf{Entertainment} & \textbf{Politics} & \textbf{Sport} & \textbf{Technology} \\
\midrule
Mistral (Unoptimized) & 0.8953 & 0.8966 & 0.8951 & 0.9733 & 0.8285 \\
GPT-3.5 (Unoptimized) & 0.9022 & 0.9337 & 0.8822 & 0.9731 & 0.9020 \\
Mistral (Optimized) & 0.9413 & 0.9681 & 0.9379 & 0.9903 & 0.9497 \\
GPT-3.5 (Optimized) & 0.9399 & 0.9413 & 0.9351 & 0.9893 & 0.9215 \\
CoT Prompting (87 mismatches) & 0.9441 & 0.9653 & 0.9444 & 0.9922 & 0.9492 \\
Judge Prompting (87 mismatches) & 0.9471 & 0.9615 & 0.9420 & 0.9883 & 0.9419 \\
\bottomrule
\end{tabular}
\label{table:Topic_0shot}
\end{table}

\begin{table}[H]
\centering
\addtocounter{Dtable}{1}
\caption{BBC Classification Experiment: F1 Scores for 5-Shot Prompting}
\begin{tabular}{lccccc}
\toprule
\textbf{Approach} & \textbf{Business} & \textbf{Entertainment} & \textbf{Politics} & \textbf{Sport} & \textbf{Technology} \\
\midrule 
Mistral (Unoptimized) & 0.9274 & 0.9526 & 0.9298 & 0.9893 & 0.9268 \\
GPT-3.5 (Unoptimized) & 0.9261 & 0.9335 & 0.9335 & 0.9902 & 0.9243 \\
Mistral (Optimized) & 0.9403 & 0.9681 & 0.9392 & 0.9913 & 0.9485 \\
GPT-3.5 (Optimized) & 0.9370 & 0.9415 & 0.9329 & 0.9903 & 0.9183 \\
CoT Prompting (84 mismatches)  & 0.9460 & 0.9641 & 0.9434 & 0.9932 & 0.9479 \\
Judge Prompting (84 mismatches) & 0.9451 & 0.9759 & 0.9522 & 0.9971 & 0.9511 \\
\bottomrule
\end{tabular}
\label{table:Topic_5shot}
\end{table}

\begin{table}[H]
\centering
\addtocounter{Dtable}{1}
\caption{Campaign Ads Experiment: Optimized Class Metrics for GPT-3.5 Predictions (0-Shot)}
\begin{tabular}{lccc}
\toprule
\textbf{Class} & \textbf{Precision} & \textbf{Recall} & \textbf{F1 Score} \\
\midrule
\textbf{attack} & 0.3157 & 0.9719 & 0.4766 \\
\textbf{contrast} & 0.5098 & 0.1741 & 0.2596 \\
\textbf{promote} & 0.9352 & 0.9056 & 0.9202 \\
\bottomrule
\end{tabular}
\label{tab:classification_metrics}
\end{table}

\begin{table}[H]
\centering
\addtocounter{Dtable}{1}
\caption{Campaign Ads Experiment: Optimized Class Metrics for Mistral Predictions (0-Shot)}
\begin{tabular}{lccc}
\toprule
\textbf{Class} & \textbf{Precision} & \textbf{Recall} & \textbf{F1 Score} \\
\midrule
\textbf{attack} & 0.4319 & 0.9438 & 0.5926 \\
\textbf{contrast} & 0.6910 & 0.4442 & 0.5408 \\
\textbf{promote} & 0.9410 & 0.9208 & 0.9308 \\
\bottomrule
\end{tabular}
\label{tab:classification_metrics}
\end{table}

\begin{table}[H]
\centering
\addtocounter{Dtable}{1}
\caption{Campaign Ads Experiment: Optimized Performance Metrics for Mistral Predictions(6-Shot)}
\begin{tabular}{lccc}
\toprule
\textbf{Class} & \textbf{Precision} & \textbf{Recall} & \textbf{F1 Score} \\
\midrule
\textbf{attack} & 0.4620 & 0.9213 & 0.6154 \\
\textbf{contrast} & 0.6257 & 0.5223 & 0.5693 \\
\textbf{promote} & 0.9498 & 0.9086 & 0.9287 \\
\bottomrule
\end{tabular}
\label{tab:classification_metrics_updated}
\end{table}

\begin{table}[H]
\centering
\addtocounter{Dtable}{1}
\caption{Campaign Ads Experiment: Optimized Performance Metrics for GPT-3.5 Predictions(6-Shot)}
\begin{tabular}{lccc}
\toprule
\textbf{Class} & \textbf{Precision} & \textbf{Recall} & \textbf{F1 Score} \\
\midrule
\textbf{attack} & 0.3012 & 0.9663 & 0.4593 \\
\textbf{contrast} & 0.4085 & 0.2143 & 0.2811 \\
\textbf{promote} & 0.9444 & 0.8728 & 0.9072 \\
\bottomrule
\end{tabular}
\label{tab:classification_metrics_updated_2}
\end{table}

\begin{table}[H]
\centering
\addtocounter{Dtable}{1}
\caption{CoT (328 Mismatches): Chain-of-Thought Performance Metrics for GPT-3.5 and Mistral (0-Shot)}
\begin{tabular}{lccc}
\toprule
\textbf{Class} & \textbf{Precision} & \textbf{Recall} & \textbf{F1 Score} \\
\midrule
\textbf{attack} & 0.4135 & 0.9270 & 0.5719 \\ 
\textbf{contrast} & 0.6591 & 0.4552 & 0.5385 \\
\textbf{promote} & 0.9453 & 0.9122 & 0.9285 \\
\bottomrule
\end{tabular}
\label{tab:classification_metrics}
\end{table}

\begin{table}[H]
\centering
\addtocounter{Dtable}{1}
\caption{JUDGE (328 Mismatches): Chain-of-Thought Performance Metrics for Mistral and GPT-3.5 Predictions (0-Shot)}
\begin{tabular}{lccc}
\toprule
\textbf{Class} & \textbf{Precision} & \textbf{Recall} & \textbf{F1 Score} \\
\midrule
\textbf{promote}  & 0.9404 & 0.9242 & 0.9323 \\
\textbf{contrast} & 0.7323 & 0.3318 & 0.4567 \\
\textbf{attack}   & 0.3755 & 0.9409 & 0.5368 \\
\bottomrule
\end{tabular}
\label{tab:class_metrics}
\end{table}

\begin{table}[H]
\centering
\addtocounter{Dtable}{1}
\caption{CoT (424 Mismatches): Chain-of-Thought Performance Metrics for GPT-3.5 and Mistral Predictions(6-Shot)}
\begin{tabular}{lccc}
\toprule
\textbf{Class} & \textbf{Precision} & \textbf{Recall} & \textbf{F1 Score} \\
\midrule
\textbf{attack} & 0.4009 & 0.9551 & 0.5648 \\
\textbf{contrast} & 0.6403 & 0.4350 & 0.5180 \\
\textbf{promote} & 0.9470 & 0.9058 & 0.9260 \\
\bottomrule
\end{tabular}
\label{tab:classification_metrics_updated_3}
\end{table}

\begin{table}[H]
\centering
\addtocounter{Dtable}{1}
\caption{Judge (424 Mismatches): Post-process Performance Metrics for GPT-3.5 and Mistral Predictions(6-Shot)}
\begin{tabular}{lccc}
\toprule
\textbf{Class} & \textbf{Precision} & \textbf{Recall} & \textbf{F1 Score} \\
\midrule
\textbf{attack} & 0.4191 & 0.9716 & 0.5856 \\
\textbf{contrast} & 0.6811 & 0.3923 & 0.4978 \\
\textbf{promote} & 0.9404 & 0.9225 & 0.9314 \\
\bottomrule
\end{tabular}
\label{tab:class_metrics}
\end{table}

\include{references}

\printbibliography

\end{document}